\documentclass[10pt,twocolumn,letterpaper]{article}

\usepackage[pagenumbers]{cvpr} 

\usepackage{graphicx}
\usepackage{amsmath}
\usepackage{amssymb}
\usepackage{dsfont}
\usepackage{booktabs}  
\usepackage{bbding}    
\usepackage{multirow,enumitem}

\newcommand{\comment}[1]{}

\usepackage{xcolor}
\usepackage[pagebackref,breaklinks,colorlinks,bookmarks=false,urlcolor=black,citecolor={green!60!black}]{hyperref}
\usepackage{soul}

\makeatletter
\def\ps@myheadings{%
    \let\@oddfoot\@empty\let\@evenfoot\@empty
    \def\@evenhead{\thepage\hfil\slshape\leftmark}%
    \def\@oddhead{{\slshape\rightmark}\hfil\thepage}%
    \let\@mkboth\@gobbletwo
    \let\sectionmark\@gobble
    \let\subsectionmark\@gobble
    }
  \if@titlepage
  \renewcommand\maketitle{\begin{titlepage}%
  \let\footnotesize\small
  \let\footnoterule\relax
  \let \footnote \thanks
  \null\vfil
  \vskip 60\p@
  \begin{center}%
    {\LARGE \@title \par}%
    \vskip 3em%
    {\large
     \lineskip .75em%
      \begin{tabular}[t]{c}%
        \@author
      \end{tabular}\par}%
      \vskip 1.5em%
    {\large \@date \par}
  \end{center}\par
  \@thanks
  \vfil\null
  \end{titlepage}%
  \setcounter{footnote}{0}%
}
\else
\renewcommand\maketitle{\par
  \begingroup
    \renewcommand\thefootnote{\@fnsymbol\c@footnote}%
    \def\@makefnmark{\rlap{\@textsuperscript{\normalfont\color{black}\@thefnmark}}}%
    \long\def\@makefntext##1{\parindent 1em\noindent
            \hb@xt@1.8em{%
                \hss\@textsuperscript{\normalfont\@thefnmark}}##1}%
    \if@twocolumn
      \ifnum \col@number=\@ne
        \@maketitle
      \else
        \twocolumn[\@maketitle]%
      \fi
    \else
      \newpage
      \global\@topnum\z@   
      \@maketitle
    \fi
    \thispagestyle{plain}\@thanks
  \endgroup
  \setcounter{footnote}{0}%
}
\makeatother

\makeatletter
\newcommand\fs@nobottomruled{\def\@fs@cfont{\bfseries}\let\@fs@capt\floatc@ruled
  \def\@fs@pre{}
  \def\@fs@post{}
  \def\@fs@mid{\kern2pt\hrule\kern2pt}%
  \let\@fs@iftopcapt\iftrue}
\makeatother

\usepackage{multibib}
\newcites{latex}{References}

\usepackage[capitalize]{cleveref}
\crefname{section}{Sec.}{Secs.}
\Crefname{section}{Section}{Sections}
\Crefname{table}{Table}{Tables}
\crefname{table}{Tab.}{Tabs.}

\def\cvprPaperID{4145} 
\def\confName{CVPR}
\def\confYear{2022}

\begin{document}

\title{Few-shot Keypoint Detection with Uncertainty Learning for Unseen Species}

\iftrue
\author{Changsheng Lu$^{\dagger}$, Piotr Koniusz\thanks{The corresponding author.$\quad$Accepted by CVPR 2022.}$\;^{,\S, \dagger}$\\
  $^{\dagger}$The Australian National University \quad 
   $^\S$Data61/CSIRO\\
{\tt\small ChangshengLuu@gmail.com, firstname.lastname@anu.edu.au}
}
\fi

\maketitle

\begin{abstract}


Current non-rigid object keypoint detectors perform well on a chosen kind of species and body parts,  
and require a large amount of labelled keypoints for training. Moreover, their heatmaps, tailored to specific body parts, cannot recognize novel keypoints (keypoints not labelled for training) on unseen species. We raise an interesting yet challenging question: how to detect both base (annotated for training) and novel keypoints for unseen species given a few annotated samples? Thus, we propose a versatile Few-shot Keypoint Detection (FSKD) pipeline, 
which can detect a varying number of keypoints of different kinds. Our FSKD 
provides the uncertainty estimation of predicted keypoints. Specifically, FSKD involves main and auxiliary keypoint representation learning, similarity learning, and keypoint localization with uncertainty modeling to tackle the localization noise. 
Moreover, we model the uncertainty across groups of keypoints by 
multivariate Gaussian distribution 
to exploit implicit correlations between neighboring keypoints. 
We show the effectiveness of our FSKD on (i) novel keypoint detection for unseen species,  (ii) few-shot Fine-Grained Visual Recognition (FGVR) and (iii) Semantic Alignment (SA) downstream tasks. For FGVR, detected keypoints improve the classification accuracy. For SA, we showcase a novel thin-plate-spline warping that uses estimated keypoint uncertainty under imperfect keypoint corespondences. 

\end{abstract}

\section{Introduction}\label{sec:introduction}
\begin{figure}[!tb]
    \centering
    \includegraphics[width=0.9\hsize]{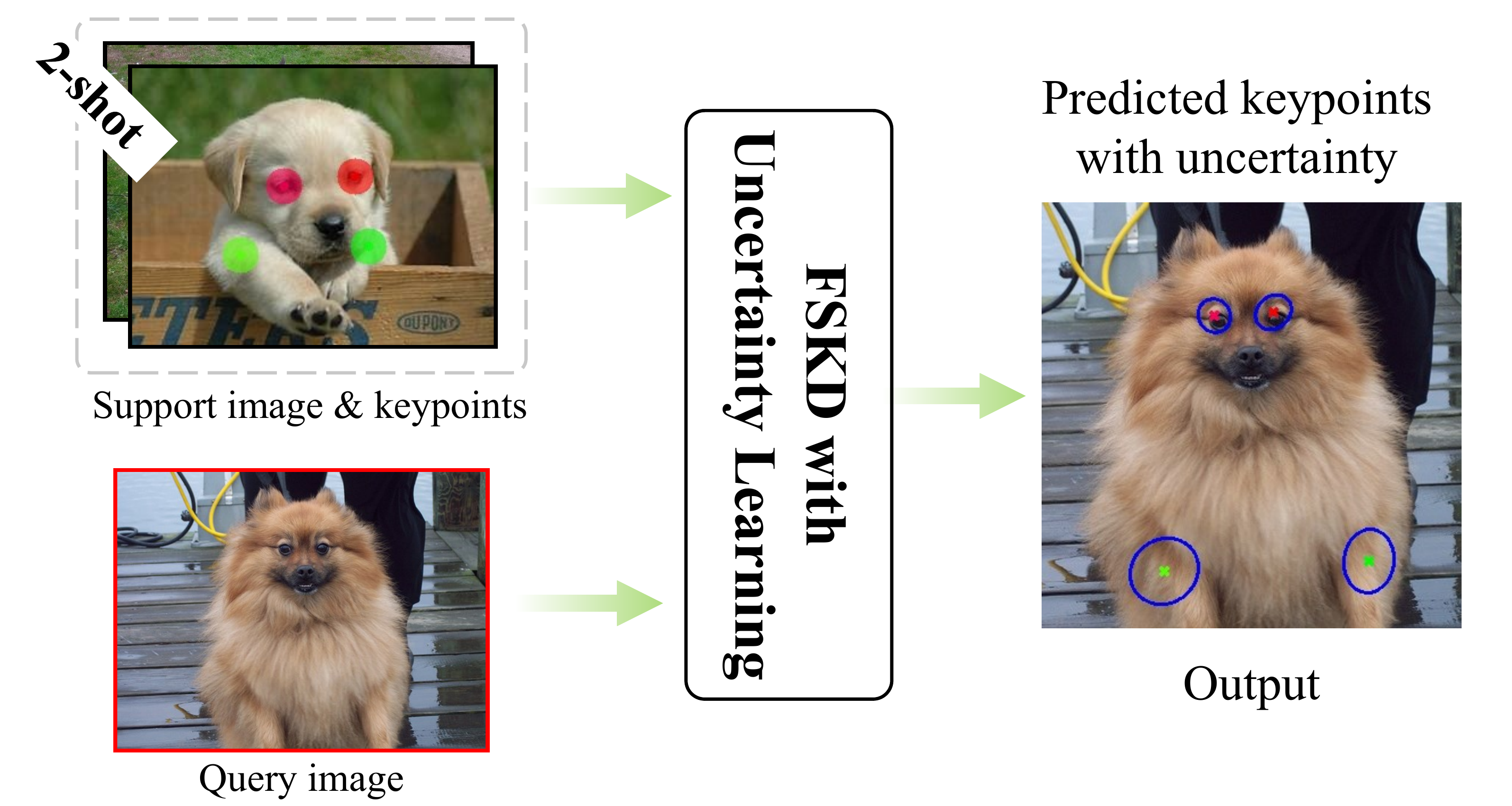}
    \caption{Illustration of few-shot keypoint detection (4-way, 2-shot protocol) with uncertainty learning. Our FSKD model successfully detects four novel keypoints and estimates the localization uncertainty given two annotated instances in unseen species. 
    }
    \label{fig:fskd-demo}
    \vspace{-10pt}
\end{figure}
\begin{figure*}[!t]
  \centering
  \begin{minipage}{0.28\linewidth}
    \centering
    \includegraphics[width=0.93\linewidth]{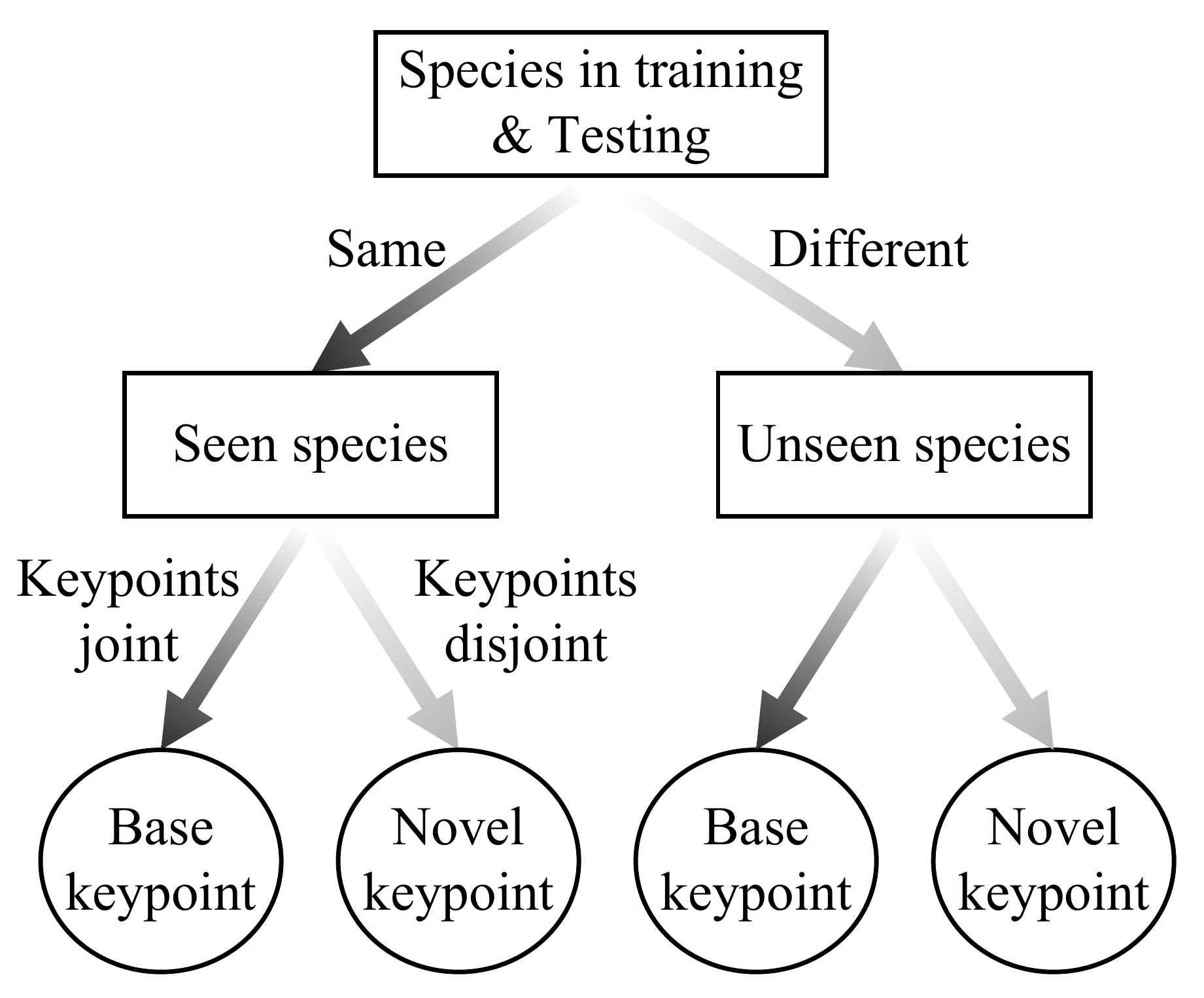}
    \caption{Different settings of FSKD lead to different levels of domain shift.}
    \label{fig:taxonomy}
  \end{minipage}
  \hspace{0.2cm}
  \hfill
  \begin{minipage}{0.69\linewidth}
    \centering
    \includegraphics[width=1.0\linewidth]{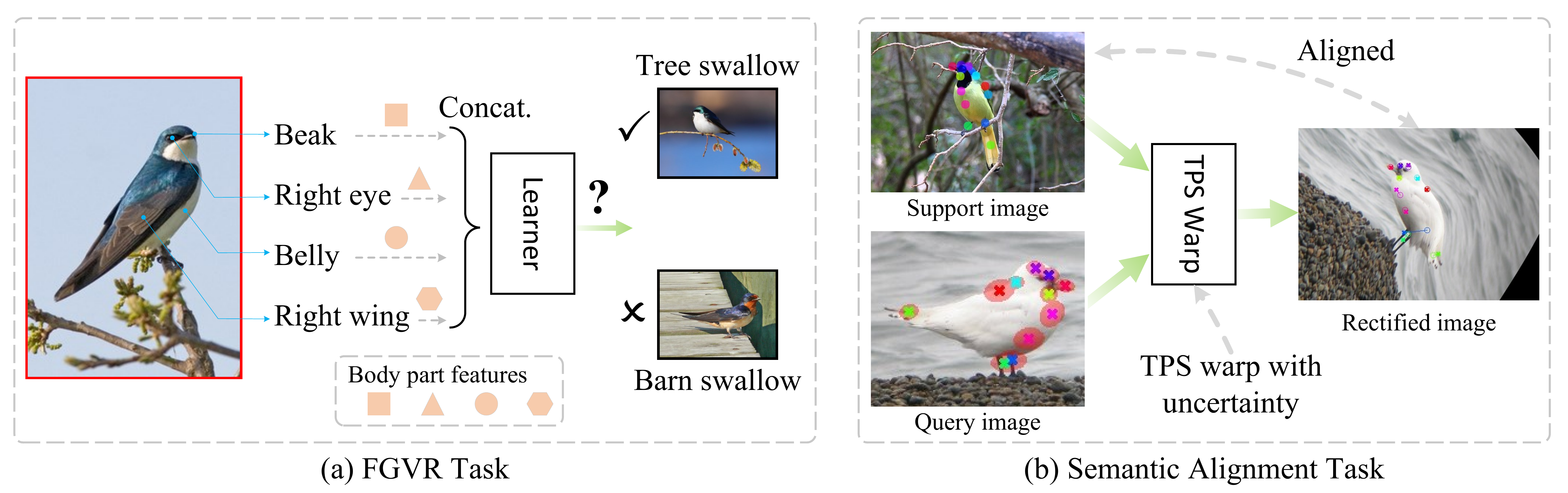}
    \caption{Downstream tasks on which we evaluate the proposed FSKD. (a) Few-shot Fine-Grained Visual Recognition (FGVR); (b) Semantic Alignment (SA).}
    \label{fig:fskd-downsteam-tasks}
  \end{minipage}
  \vspace{-10pt}
\end{figure*}

Deep fully-supervised pose estimation has proven its excellence in detecting keypoints on humans \cite{cao2019openpose,fang2017rmpe,he2017mask,newell2016stacked,toshev2014deeppose}. However, these keypoint detectors are dedicated to specific species and body parts on which they are trained. They are not reusable for unseen species, and consume large amounts of annotated data. In contrast, given a few samples, a child can adequately recognize and generalize a keypoint on a paw of rabbit, cat, dog, kangaroo under varying poses \eg, jumping, crouching, or walking. 
%
By comparison, machine perception is inferior to biological perception \cite{wang2020generalizing}. 
Inspired by the progress in few-shot learning \cite{koch2015siamese,vinyals2016matching,sung2018learning,deeper_look2,Simon_2022_WACV,haohao_CVPR}, we propose Few-shot Keypoint Detection (FSKD) which can learn from 
few keypoints 
and recognise previously unseen keypoint types even for species unseen during training (Fig.~\ref{fig:fskd-demo}).  


As keypoints provide crucial structural and semantic information, FSKD has numerous promising applications such as semi-automatic labelling, face alignment \cite{kowalski2017deep,sun2013deep}, Fine-Grained Visual Recognition (FGVR) \cite{tang2020revisiting}, animal behavior analysis \cite{pereira2019fast}, \etc. The use of keypoints can also simplify the labor-intensive surveillance of wildlife \cite{marstaller2019deepbees, zhang2020omni}. 


In this paper, we propose a versatile FSKD approach which copes with various levels of domain shift. Fig.~\ref{fig:taxonomy} shows that \emph{the categories of species used for training and testing may be the same or different (top branches), and also the keypoint types of specific body parts may be the same or disjoint (bottom branches)}. The easiest problem setting assumes the same kind of species and the same types of keypoints throughout training and testing. However, if the species as well as keypoint types used for training and testing are both disjoint, the problem is challenging due to high levels of domain shift.


We note that learning and generalizing based on a few of samples is hard due to the limited number of annotations and a large variability of samples. 
Moreover, there exist large amounts of interfering noise and similar local patterns in an image, 
which can  challenge  FSKD and deem it arguably a much harder problem than Few-shot Object Detection (FSOD) \cite{kang2019few,fan2020few,Zhang_2022_CVPR}. In contrast to objects which usually have well-defined boundaries, keypoints lack the clear boundary and correspond to some local regions rather than exact coordinates. Consequently, keypoints inherently exhibit ambiguity and location uncertainty, which are reflected in both groundtruth keypoints and predictions. 
Thus, we develop an FSKD approach 
which can deal with domain shifts and model the uncertainty of localization.

Our FSKD firstly extracts the deep representations of support keypoints to build the keypoint prototypes (one per keypoint type), which are correlated against the query feature map to yield keypoint-specific attentive features. 
After descriptor extraction, attentive features are transformed into descriptors which will be used for subsequent keypoint localization. 
To address the limited types of base training keypoints, we introduce the generated auxiliary keypoints into learning. Though these keypoints show poor matching between support and query images, they boost the keypoint diversity and significantly help infer novel keypoints. The difficulty of FSKD results in imperfect keypoint predictions over novel keypoints. To compensate for this effect and deal with inherent ambiguity and noise of keypoints, we propose to model the localization uncertainty by learning the covariance for individual or multiple keyoints, which allows  larger tolerance of those noisy keypoints by the loss function. 
%
Our contributions  are summarized as follows:
\renewcommand{\labelenumi}{\roman{enumi}.}
\vspace{-0.1cm}
\hspace{-1.0cm}
\begin{enumerate}[leftmargin=0.6cm]
    \item We propose a flexible few-shot keypoint detection (FSKD) model that can detect varying types and numbers of keypoints given one or more annotated samples. 
    \item Both localization and semantic uncertainty are modeled within our localization networks, where an uncertainty-aided grid-based locator (UC-GBL) is proposed. Moreover, we propose the multi-scale UC-GBL in order to reduce the risk of mislocalization. 
    \item We employ the low-quality auxiliary keypoints during learning, and model the covariance for coupled main and auxiliary keypoints to improve generalization.
    \item Convincing experiments show that our FSKD can detect novel keypoints on unseen species. With a simple modification, we extend our FSKD to act as a simple keypoint detector applied to few-shot fine-grained visual recognition and semantic alignment (Fig.~\ref{fig:fskd-downsteam-tasks}).
\end{enumerate}

To our best knowledge, our work is the first attempt to model keypoint detection as few-shot learning. 

\section{Related Work}\label{sec:related-work}
%
\noindent\textbf{Few-shot Learning (FSL):}
Initially, FSL was dedicated to image classification based on few samples \cite{fei2006one,wang2020generalizing}. 
The current  FSL pipelines are based on deep backbones and mainly focus on i) metric learning \cite{koch2015siamese,vinyals2016matching,snell2017prototypical,sung2018learning}; ii) optimization, \eg, MAML \cite{finn2017model} rapidly adapts to new tasks; and iii) data hallucination \cite{zhang2019few,zhang2021hallucination}. 
FSL 
has expanded into other computer vision tasks including few-shot  segmentation \cite{li2020fss, li2021adaptive} and object detection \cite{fan2020few, kang2019few, zhang2021hallucination,Zhang_2022_CVPR}.

\comment{
Few-shot learning aims to learn and generalize from a few samples to numerous ones, which is very chellenging in the fact of large variability among instances. Currently, the majority of few-shot learning algorithms mainly focus on image classification \cite{wang2020generalizing}.
An early attempt is to build Bayesian probabilistic model on visual categories whose posterior will be updated when coming new observations \cite{fei2006one}. Since traditional model is shallow, current approaches \cite{koch2015siamese,vinyals2016matching,snell2017prototypical,sung2018learning,finn2017model} are dominantly based on deep networks in order to obtain feature representation with higher expressiveness. Koch \etal \cite{koch2015siamese} firstly apply Siamese networks to few-shot learning and perform simple metric learning using deep features. The key insight is learning the similarity of categories in metric space and enforcing samples from same category to own higher similarity. Matching networks \cite{vinyals2016matching} formalize $N$-way $K$-shot learning protocol, where similarity is computed by matching each query image and support images. Prototypical networks \cite{snell2017prototypical} produce prototypes by averaging the support samples, and each prototype represents a class cluster. Thus the classification for each query image could be applied by assigning the softmax over the distance to prototypes. Considering above deep methods all employ non-parametric distance metric, relation networks \cite{sung2018learning} automate comparator by adopting a network to learn similarity. In contrast to explicit metric learning, MAML \cite{finn2017model} aims at learning a meta-model, which could be fastly adapted to a new task through one or several gradient steps of finetuning. The advantage of MAML is easy to be used in any learning model, namely model-agnostic, but the training is usually hard and not stable. In this paper, we extends few-shot learning from image classification to keypoint detection, where most of terms and settings are same but sligtly different in the $N$-way learning. For example, in few-shot classification $N$-way means $N$ image classes while ours represents $N$ keypoint categories and we allow part of keypoints are invisible.

CNN based object detector has achieved great success \cite{ren2015faster, he2017mask, redmon2016you, redmon2018yolov3}, but it relies on massive human effors for bounding box annotation. In order to reduce the dependence on labels, some researchers turn to few-shot object detection (FSOD) \cite{fan2020few, kang2019few}. Current trend on FSOD mainly builds upon proposal-free (\eg, RCNN series) or proposal-based (\eg, YOLO series) object detectors, and focuses on three perspectives: learning better feature via metric learning \cite{kang2019few, fan2020few}, improving proposal generation using attention mechanism \cite{fan2020few}, and increasing data variability through hallucination \cite{zhang2021hallucination}. Our work has close relation to FSOD since keypoints can be regarded as local patchs and thus keypoint detection can be modified into object detection problem. However, there are essential distinctions that make FSOD inappropriate to handle FSKD. The first is that holistic object always has clear boundary and visually more salient from background while the local body part (\eg, knee and nape) lacks delimitation and inherently has location uncertainty, which needs to be properly modeled. In addition, the scale, local appearance ambiguity, and unnecessary bounding box annotation for body parts are the other issues.

Few-shot learning so far has covered most computer vision tasks, including image classification, object detection, and segmentation \cite{li2020fss, li2021adaptive}, while to our best knowledge, this is the first time of few-shot learning meets keypoint detection, which will be formally introduced and targeted in this paper.
}

\noindent\textbf{Keypoint Detection:}
Compared to traditional keypoint detectors \cite{lowe2004sift,derpanis2004harris}, the deep learning methods are more general and can be categorized into two kinds, where the first kind uses the heatmap regression followed by post-processing to search the keypoint with maximal heat value \cite{sun2013deep,newell2016stacked,cao2019openpose,fang2017rmpe,he2017mask}, and the second kind is directly performing regression on the keypoint position \cite{carreira2016human,toshev2014deeppose}, which is also adopted in our work. As for the heatmap regression based approaches, they can be further divided into top-down \cite{sun2019deep,fang2017rmpe,he2017mask} and bottom-up pose estimators \cite{newell2016stacked,cao2019openpose,cheng2020higherhrnet}. 
Recently, some works \cite{cao2019cross, li2020deformation} perform cross-domain adaptation or shape deformation by leveraging the large-scale source datasets. Novotny \etal \cite{novotny2018self} use self-supervision to learn matching features between a pair of transformed images. 
Though these approaches can limit annotation burden when learning, they cannot directly detect novel keypoints on unseen species based on a few samples.

\comment{
Traditional methods exploit low-level features to discover corner points \cite{lowe2004sift,derpanis2004harris} but are mostly limited to detect stable visual primitives across scale-space.  
In contrast, recent keypoint detectors based on  deep learning  
can be categorized into two families, namely (i) the heatmap-based models with a post-processing search for maximal heat value indicating keypoint \cite{sun2013deep,newell2016stacked,cao2019openpose,fang2017rmpe,he2017mask}, and (ii) keypoint positional regression models \cite{carreira2016human,toshev2014deeppose}.  Moreover, heatmap-based approaches can be divided into top-down \cite{sun2019deep,fang2017rmpe,he2017mask} and bottom-up pose estimators \cite{newell2016stacked,cao2019openpose,cheng2020higherhrnet}. 
Recently, some works \cite{cao2019cross, li2020deformation}  perform cross-domain adaptation or shape deformation by leveraging the large-scale source datasets. Novotny \etal \cite{novotny2018self} use self-supervision to learn matching features between a pair of transformed images with different views. 
Domain adaptation and self-supervision help given limited  annotations but they cannot directly detect novel keypoints on unseen species based on a few samples.
We adopt the keypoint positional regression.
}
\comment{
Traditional approaches adopt low-level signals such as image gradients to discover corner points \cite{lowe2004sift,derpanis2004harris}. These methods are unsupervised and easy-used while also computationally time-consuming or not robust enough. In additon, we usually intend to detect meaningful keypoints like joints or body parts on animals, which would be preferablly suitable for learning models. There are two mainstreams deep-learning keypoint detection approaches. The first is regressing heatmap followed by post-processing to search maximal heat value as keypoint \cite{sun2013deep,newell2016stacked,cao2019openpose,fang2017rmpe,he2017mask}. And the second stream is directly regressing keypoint position \cite{carreira2016human,toshev2014deeppose}, which is also adopted in our work. For the first stream of approach, namely, heatmap regression, bottom-up human pose estimator \cite{newell2016stacked,cao2019openpose,cheng2020higherhrnet} or the top-down one \cite{sun2019deep,fang2017rmpe,he2017mask} are proposed, which differs on regressing heatmap first or detecting person first. Though existing human pose estimator is quite advanced, the main concern is the large consumption of mannual keypoint annotations and it might easily underperform when applying the model trained on seen species to novel species that has different anatomy. In order to bridge the gap between species, some works \cite{cao2019cross, li2020deformation} try to perform cross-domain adaptation or shape deformation where the large labelled source datasets or synthetic datasets are leverage. In addition, the self-supervision could be also exploited to relieve annotation burden by using a pair of geometrically transformed images \cite{novotny2018self}. Nevertheless, most current approaches are dedicatedly designed to solve the base keypoint detection on seen species, while our target is towards detecting novel keypoints on unseen species.

}

\noindent\textbf{Uncertainty in Computer Vision:}
%
The uncertainty generally consists of \emph{aleatoric uncertainty} and \emph{epistemic uncertainty} \cite{kendall2017uncertainties,blundell2015weight} modeled by a Gaussian over the predictions and placing a distribution over the model weights (\ie, Bayesian Neural Network \cite{blundell2015weight}), respectively. In this paper, we mainly focus on the heteroscedastic aleatoric uncertainty, popular in a number of applications. Kendall \etal \cite{kendall2018multi} use uncertainty to relatively weigh multi-task loss functions between depth regression and segmentation. He \etal \cite{he2019bounding} and Choi \etal \cite{choi2019gaussian} incorporate uncertainty into  bounding box regression of Faster R-CNN \cite{ren2015faster} and YOLOv3 \cite{redmon2018yolov3}. 
However, these models treat multiple variables independently, 
while we model uncertainty by covariance to capture underlying relations between variables. 

\comment{
Uncertainty estimation is very useful in many tasks, especially in autonomous driving, which can help deep model to show uncertain when facing noise or rare patterns, therefore, avoiding making the inappropriate decisions that may cause tragedy. Uncertainty generally consists of aleatoric uncertainty and epistemic uncertainty \cite{kendall2017uncertainties,blundell2015weight}. The former reflects the uncertainty degree on the model prediction $f(x)$ due to bad quality, noise, or disturbances in the input data $x$, and this uncertainty could be modeled by a Gaussian distribution $G(\mu, \sigma^2)$ with mean $\mu=f(x)$ for the output $y$, where $y\sim G(\mu, \sigma^2)$. While for latter one, the epistemic uncertainty indicates the ignorance of the model what has never learned and could be modeled by placing a distribution $p(w|x, y)$ over the weights $w$, of which the representative approach is Bayesian neural network (BNN) \cite{blundell2015weight}. In this paper, we mainly focus on the heteroscedastic aleatoric uncertainty in our FSKD.

A number of computer vsion applications involve aleatoric uncertainty. Kendall \etal \cite{kendall2018multi} utilize uncertainty to relatively weigh the multi-task loss functions of per-pixel depth regression and segmentation. He \etal \cite{he2019bounding} and Choi \etal \cite{choi2019gaussian} incorporate uncertainty in regressing bounding box in Faster R-CNN and YOLOv3, respectively, reducing the influence of hard examples and improving localization accuracy. Moreover, it can also be extended to 3D object bounding box regression \cite{chen2020monopair}. Recently, Luo \etal \cite{luo2021rethinking} propose adaptive heatmap regression which adjusts the standard deviation to produce pseudo keypoint heatmap as groundtruth for supervision. It needs to mention that all above approaches are based on the assumption of independence for each variable that involves uncertainty. Instead, in our work we do not have this condition and the variables can be correlated.

%
}


\section{Few-shot Keypoint Detection}\label{sec:FSKD}
\begin{figure*}[!t]
    \centering
    \includegraphics[width=.9\linewidth]{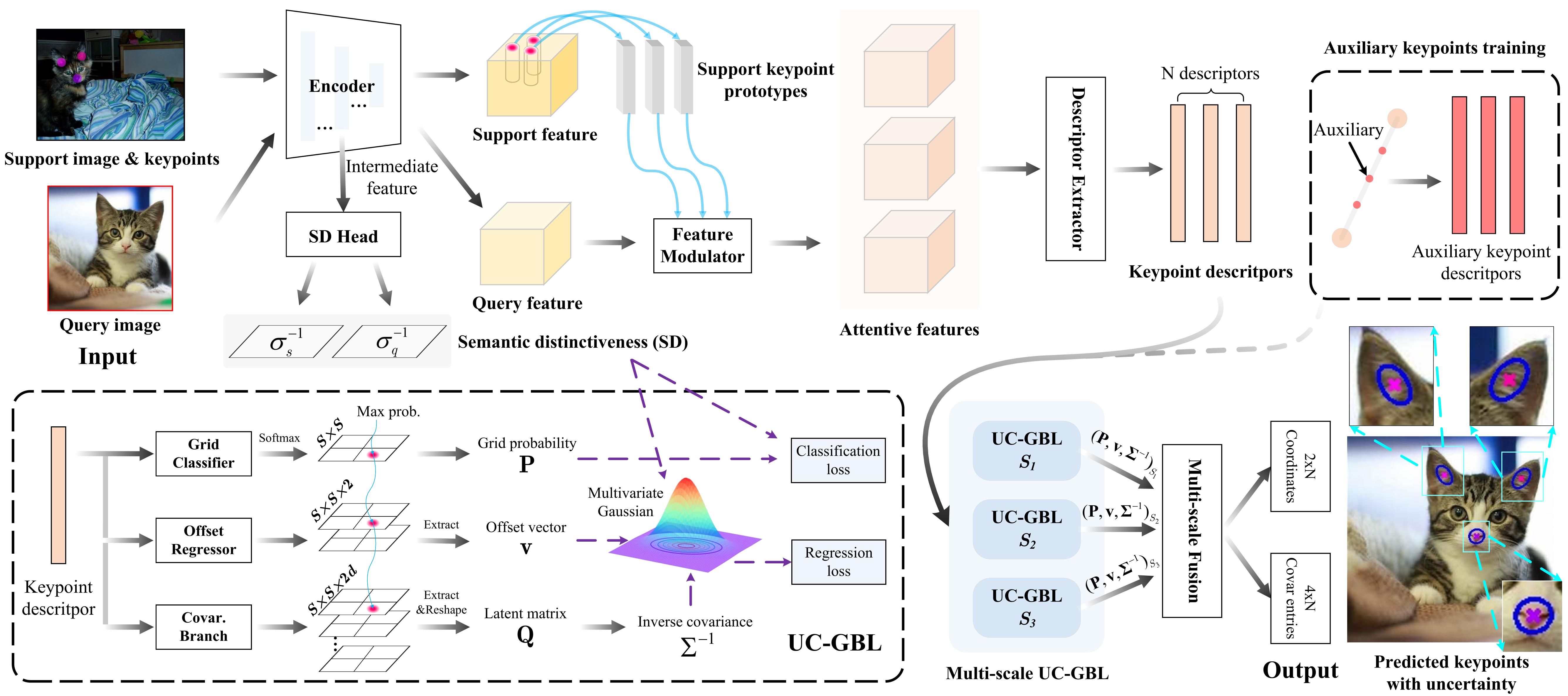}
    \caption{Few-shot keypoint detection pipeline. The whole model aims to predict the keypoints with uncertainty in a query image given the support keypoints. 
    The  prototypes guide the query feature to generate keypoint-specific attentive features, which will be condensed into descriptors for keypoint localization through multi-scale UC-GBL. Via interpolation, the auxiliary keypoints with poor matching quality are also incorporated into the learning process to boost the  generalization ability of FSKD.}
    \label{fig:fskd-pipeline}
    \vspace{-10pt}
\end{figure*}

\subsection{Architecture Overview}
Given the sampled support image and keypoints, FSKD aims to detect the corresponding keypoints in query image, where the hardest setting includes species and keypoint types that are both disjoint between training and testing. For \emph{$N$ support keypoints} and \emph{$K$ support images}, the problem is dubbed as \emph{$N$-way-$K$-shot detection}. 

An overview of our FSKD pipeline is shown in Fig.~\ref{fig:fskd-pipeline}, which includes feature encoder $\mathcal{F}$, feature modulator $\mathcal{M}$, descriptor extractor $\mathcal{P}$, and uncertainty-aided grid-based locator $\mathcal{G}$ (UC-GBL). Moreover, we also learn the semantic distinctiveness (SD) by adding a side branch $\mathcal{D}$ (SD head). In the following, we will describe each module in detail. 


\subsection{Keypoint Prototypes and Descriptors Building}
\noindent\textbf{Keypoint Embedding:}
Let support and query images $\mathbf{I}_\text{s}$ and $\mathbf{I}_\text{q}$ be mapped to $\mathcal{F}(\mathbf{I}_\text{s})$ and $\mathcal{F}(\mathbf{I}_\text{q})$ in feature space $\mathbb{R}^{l \times l \times C}$ using a weight-shared convolutional encoder $\mathcal{F}$. Given a support keypoint at location $\mathbf{u}=[x, y]^{\intercal}$, 
we extract the keypoint representation $\mathbf{\Phi} = \mathcal{A}(\mathcal{F}(\mathbf{I}_\text{s}), \mathbf{u})$ via $\mathcal{A}$. Operator $\mathcal{A}$ could be integer-based indexing \cite{novotny2018self}, bilinear interpolation \cite{jaderberg2015spatial, he2017mask}, or Gaussian pooling. 
Feature representations $\mathbf{\Phi}$ must contain some local context of keypoint to be sufficiently discriminative 
during the matching step. Therefore, we employ Gaussian pooling. 
Let $\mathbf{u}_{k, n}$ be the $n$-th keypoint in the $k$-th support image $\mathbf{I}_\text{s}^k$ and $\mathcal{F}(\mathbf{I}_\text{s}^k)(\mathbf{x})$ be the feature vector at position $\mathbf{x} \in \mathbb{R}^2$. The keypoint representation $\mathbf{\Phi}_{k, n}$ can be obtained as
\begin{equation}
  \mathbf{\Phi}_{k, n} = \sum\nolimits_{\mathbf{x}} \exp(-\parallel \mathbf{x}-\mathbf{u}_{k,n} \parallel _2^2/2\xi  ^2) \cdot \mathcal{F}(\mathbf{I}_\text{s}^k)(\mathbf{x}),
\end{equation}
where  $\xi$ is the standard deviation. Following prototypical networks \cite{snell2017prototypical}, each support keypoint prototype (SKP) $\mathbf{\Phi}_n$ can be obtained by averaging the same type of keypoint representations across support images as 
\begin{equation}
  \mathbf{\Phi}_n = \frac{1}{K} \sum\nolimits_{k=1}^{K} \mathbf{\Phi}_{k, n},
\end{equation}
where $n=1,\cdots, N$. Thus we have $N$ SKPs generated in the $N$-way $K$-shot learning process.

\noindent\textbf{Support and Query Correlation:}
In order to guide FSKD to discover the corresponding keypoints in the query image, 
SKP $\mathbf{\Phi}_n$ needs to be correlated against query feature map $\mathcal{F}(\mathbf{I}_\text{q})$. To this end, we adopt a feature modulator $\mathcal{M}$, which takes both $\mathbf{\Phi}_n$ and $\mathcal{F}(\mathbf{I}_\text{q})$ as input to produce the attentive features $\mathbf{A}_{n} = \mathcal{M}(\mathcal{F}(\mathbf{I}_\text{q}), \mathbf{\Phi}_n), n=1,2,\cdots,N$. For $\mathcal{M}$ we could choose simple correlation, spatial attention, channel attention, \etc. In this paper, we use simple correlation for its efficiency, which is given as $\mathbf{A}_{n}(\mathbf{p}) = \mathcal{F}(\mathbf{I}_\text{q})(\mathbf{p}) \odot \mathbf{\Phi}_n$, 
where $\mathbf{p} \in \mathbb{R}^2$ runs over the feature map of size $l \times l$, and $\odot$ is the channel-wise multiplication. 
Feature correlation performs the similarity learning between support and query, which `activates' the local regions in the query feature map that significantly correlate with SKP. 

\comment{
In order to guide FSKD to discover the corresponding keypoints in query image, 
the relation between SKP $\mathbf{\Phi}_n$ and query feature $\mathcal{F}(\mathbf{I}_\text{q})$ should be properly correlated. We adopt a feature modulator $\mathcal{M}$ to serve this role, which takes both $\mathbf{\Phi}_n$ and $\mathcal{F}(\mathbf{I}_\text{q})$ as input, producing the attentive features $\mathbf{A}_{n} = \mathcal{M}(\mathcal{F}(\mathbf{I}_\text{q}), \mathbf{\Phi}_n), n=1,2,\cdots,N$.  Feature modulator $\mathcal{M}$ could have various choices such as simple correlation, spatial attention, channel attention, \etc. In this paper, we select simple correlation for its efficiency, which is $\mathbf{A}_{n}(\mathbf{p}) = \mathcal{F}(\mathbf{I}_\text{q})(\mathbf{p}) \odot \mathbf{\Phi}_n$, 
where $\mathbf{p} \in \mathbb{R}^2$ runs over each position in $l \times l$, and $\odot$ refers channel-wise multiplication. It needs to mention that feature correlation essentially performs the similarity learning between support and query, which encourages the local region in query feature that has significant relation with SKP to response actively.
}


\noindent\textbf{Keypoint Descriptor Extraction:}
After performing feature correlation, we then project each attentive feature $\mathbf{A}_{n}$ into keypoint descriptor $\mathbf{\Psi}_n$ via a descriptor extractor $\mathcal{P}$ to decrease the dimensionality, namely, $\mathbf{\Psi}_n = \mathcal{P}(\mathbf{A}_{n})$. 

\comment{
Having performed the feature correlation, we then project each attentive feature $\mathbf{A}_{n}$ into keypoint descriptor $\mathbf{\Psi}_n$ via a descriptor extractor $\mathcal{P}$ to further encode it as $\mathbf{\Psi}_n = \mathcal{P}(\mathbf{A}_{n})$. 
}


\subsection{Keypoint Localization \& Uncertainty Learning}
\subsubsection{Vanilla Grid-based Locator (GBL)}
Instead of regressing global location for each keypoint descriptor $\mathbf{\Psi}_n$, we approach the keypoint localization as grid classification and local offset regression, where a grid classifier $\mathcal{G}_\text{c}$ and an offset regressor $\mathcal{G}_\text{o}$ are used. Let $\mathbf{v}\in \mathbb{R}^{2}$ be the offset to the center of a grid 
where the keypoint falls. 
The predicted grid $\mathbf{g}\!\in\!\{0,\cdots,S\!-\!1\}\!\times\!\{0,\cdots,S\!-\!1\}$ is obtained by the 2D index of the maximum in the grid probability map $\mathbf{P} \in \mathbb{R}^{S\times S}$, where $\mathbf{P}\!=\!\text{Softmax}(\mathcal{G}_\text{c}(\mathbf{\Psi}_n))$, and the predicted offset $\mathbf{v}$ can be retrieved from the vector field $\mathcal{G}_\text{o}(\mathbf{\Psi}_n) \in \mathbb{R}^{S\times S \times 2}$ at grid $\mathbf{g}$ (see Fig.~\ref{fig:fskd-pipeline}). In addition, we construct the groundtruth (GT) offset $\mathbf{v}^*$ via
\begin{equation}
  \label{eq:GBL}
    \begin{array}{ccc}
      \mathbf{t} = {\mathbf{u}^* S}/{l_0} & \mathbf{z} = \lfloor {\mathbf{t}} \rfloor + 0.5 & \mathbf{v}^* = 2(\mathbf{t} - \mathbf{z}),
    \end{array}
\end{equation}
where $\mathbf{u}^*\in\mathbb{R}^{2}$ is a GT keypoint in square-padded query image with edge length $l_0$, $\mathbf{t}\in\mathbb{R}^{2}$ is the transformed coordinate in the grid frame, and $\mathbf{z}$ is the grid center. Furthermore, the grid label can be formed  as 
$g^*=\lfloor {\mathbf{t}_y} \rfloor S + \lfloor {\mathbf{t}_x} \rfloor$, and 
$\mathbf{v}^*\in [-1,1)^{2}$. 
With $\mathbf{v}^*$ and $g^*$, we 
design a cross-entropy grid classification loss $\mathcal{L}_{\text{cls}}$, and an offset regression loss $\mathcal{L}_{\text{os}}$ using MSE to minimize the vanilla GBL $(\mathcal{G}_\text{c}, \mathcal{G}_\text{o})$.

\comment{
Instead of regressing global location for each keypoint descriptor $\mathbf{\Psi}_n$, we decompose the keypoint localization into grid classification and local offset regression, where a grid classifier $\mathcal{G}_\text{c}$ and offset regressor $\mathcal{G}_\text{o}$ are employed. The offset $\mathbf{v}$ gives the replacement to the center of grid $g$ where the keypoint falls. Assuming the grid size (or called as scale) is $S\times S$. The grid $\mathbf{g}\in\mathbb{R}^2$ is obtained by indexing the maximum from grid probability map $\mathbf{p} \in \mathbb{R}^{S\times S}$, where $\mathbf{p}=\text{Softmax}(\mathcal{G}_\text{c}(\mathbf{\Psi}_n))$. The offset $\mathbf{v}$ can be retrieved from the vector field $\mathcal{G}_\text{o}(\mathbf{\Psi}_n) \in \mathbb{R}^{S\times S \times 2}$ at grid $g$ (see Fig.~\ref{fig:fskd-pipeline}). In addition, we construct the groundtruth (GT) offset $\mathbf{v}^*$ via
\begin{equation}
  \label{eq:GBL}
    \begin{array}{ccc}
      \mathbf{t} = {\mathbf{u}^* S}/{l_0} & \mathbf{z} = \lfloor {\mathbf{t}} \rfloor + 0.5 & \mathbf{v}^* = 2(\mathbf{t} - \mathbf{z}) \\
    \end{array}
\end{equation}
where $\mathbf{u}^*$ is GT keypoint in square-padded query image with edge length $l_0$, $\mathbf{t}$ is the transformed coordinates in the grid frame, and $\mathbf{z}$ is the grid center. Further, the grid label can be achieved through $g^*=\lfloor {\mathbf{t}_y} \rfloor S + \lfloor {\mathbf{t}_x} \rfloor$. $\mathbf{v}^*$ is regularized into $-1 \sim 1$. With the constructed supervised signal $\mathbf{v}^*$ and $g^*$, we are able to build cross-entropy grid classification loss $\mathcal{L}_{\text{cls}}$, and offset regression loss $\mathcal{L}_{\text{os}}$ using MSE to optimize the vanilla GBL $\{\mathcal{G}_\text{c}, \mathcal{G}_\text{o}\}$.
}

\subsubsection{Localization and Semantic Uncertainty}\label{sec:loc-sem-uc}


Unlike approaches \cite{he2019bounding, choi2019gaussian, luo2021rethinking, kendall2018multi}, we use covariance $\mathbf{\Sigma}$ to model the localization uncertainty of individual or multiple keypoints. Let $\mathcal{N}(\mathbf{x} ; \mathbf{v}^*, \mathbf{\Sigma})$  be the multivariate Gaussian distribution with
$\mathbf{x}, \mathbf{v^{*}} \in \mathbb{R}^{k}, k\geq2$, and $\mathbf{\Sigma} \in \mathbb{S}^{k}_{++}$. Let $\mathbf{x}$ be the predicted offset for a keypoint (or stacked multiple keypoints) and $\mathbf{v^{*}}$ be the GT, then one may write the negative log-likelihood (NLL) loss as 
\begin{equation}\label{eq:nll-gaussian-loss}
  \begin{aligned}
      &\mathcal{L}_{\text{os-nll}} = -\mathbb{E}\log \mathcal{N}(\mathbf{x} ; \mathbf{v}^*, \mathbf{\Sigma}) \\
      &\equiv \frac{1}{2}\mathbb{E}[(\mathbf{x} - \mathbf{v}^*)^{\intercal} \mathbf{\Sigma}^{-1} (\mathbf{x} - \mathbf{v}^*) + \log\det(\mathbf{\Sigma})].
  \end{aligned}
\end{equation}
However, $\mathcal{L}_{\text{os-nll}}$ in Eq.~\ref{eq:nll-gaussian-loss} relies on the computation of the inverse of covariance matrix $\mathbf{\Sigma}$, which is costly and unstable in back-propagation 
especially when $k\geq 4$. 
Thus, 
we replace $\mathbf{\Sigma}$ with the precision matrix $\mathbf{\Omega} = \mathbf{\Sigma}^{-1}$, and obtain 
\begin{equation}\label{eq:nll-gaussian-loss2}
  \mathcal{L}_{\text{os-nll}} = \frac{1}{2}\mathbb{E}[(\mathbf{x} - \mathbf{v}^*)^{\intercal} \mathbf{\Omega} (\mathbf{x} - \mathbf{v}^*) - \log\det(\mathbf{\Omega})].
\end{equation}
As a result, the $\mathcal{L}_{\text{os-nll}}$ can be easily computed as long as $\mathbf{\Omega}\succ\boldsymbol{0}$. 
To guarantee this, we let $\mathbf{\Omega} = \frac{1}{d}\mathbf{Q}\mathbf{Q}^{\intercal}$, 
where $\mathbf{Q} \in \mathbb{R}^{k\times d}$ ($d\ge k$) is the latent matrix learned from our covariance branch network. In extreme case, 
a small $\epsilon \rightarrow 0$ can be added to ensure $\det(\mathbf{\Omega}) > 0$.

Firstly, we investigate learning the covariance per keypoint by adding a covariance branch $\mathcal{G}_\text{v}$ to GBL (Fig.~\ref{fig:fskd-pipeline}), whose output is the latent covariance field $\mathcal{G}_\text{v}(\mathbf{\Psi}_n) \in \mathbb{R}^{S \times S \times 2d}$. 
Then $\mathbf{Q} \in \mathbb{R}^{2 \times d}$ which encodes the covariance information for a given keypoint can be extracted from grid $\mathbf{g}$. Secondly, we investigate learning relations for multiple keypoints within group (\ie, $m$ keypoints per group) by a multi-keypoint covariance branch $\mathcal{G}_\text{mkv}$ whose output is $\mathbf{Q}_\text{mkv} \in \mathbb{R}^{2m \times 2m}$. 

In addition to $\mathbf{\Sigma}$ which reflects the localization uncertainty over keypoints, following \cite{kendall2017uncertainties, novotny2018self}, we also model the semantic uncertainty $\boldsymbol{\sigma}$ 
by learning a single-channel SD map $\boldsymbol{\sigma}^{-1}_s \in \mathbb{R}^{H \times W \times 1}$  via SD head $\mathcal{D}$  (Fig.~\ref{fig:fskd-pipeline}) whose values are in range $(0,1)$. The higher the value, the more distinctive perceptually a keypoint is. Let $\boldsymbol{\sigma}^{-1}_s$ and $\boldsymbol{\sigma}^{-1}_q$ be the support and query SD maps. Thus, 
for each keypoint descriptor $\mathbf{\Psi}_n$, we extract the corresponding SD scalar as $w_n = \frac{1}{2}(\sigma_{\text{s},\mathbf{u}_n}^{-1} +\sigma_{\text{q},\mathbf{u}'_n}^{-1})$, where $\sigma_{\text{s},\mathbf{u}_n}^{-1}$ and $\sigma_{\text{q},\mathbf{u}'_n}^{-1}$ are  support and query  SD map values at keypoint locations  $\mathbf{u}_n$ and $\mathbf{u}'_n$. Let $\mathbf{W}=diag([w_1, w_1, \cdots, w_m, w_m])\in \mathbb{S}^{2m}_{++}$ be a diagonal matrix with entries $w_n$. We introduce $\mathbf{W}$ into Eq.~\ref{eq:nll-gaussian-loss2}:
\begin{equation}\label{eq:dev-uc-loss}
   \mathcal{L}_\text{uc}=\frac{1}{2}\mathbb{E}[(\mathbf{x} - \mathbf{v}^*)^{\intercal} (\mathbf{\Omega} + \beta\mathbf{W}) (\mathbf{x} - \mathbf{v}^*) - \log(\det(\mathbf{\Omega}\mathbf{W}^\beta))],
\end{equation}
where $\beta$ is a trade-off (we set $\beta\!=\!1$). We also use $w_n$ to reweight the cross-entropy grid classification loss $\mathcal{L}_{\text{cls-uc}}=-\mathbb{E}[\sqrt{w_n}\,\langle\mathds{I}(g^*),\log(\text{vec}(\mathbf{P}))\rangle]$ where $\text{vec}(\cdot)$ vectorizes a matrix,  $\mathds{I}(\cdot)$ is a one-hot encoding of scalar. Both $\mathcal{L}_\text{uc}$ and $\mathcal{L}_{\text{cls-uc}}$ 
are used by the uncertainty-aided GBL (UC-GBL).

Compared to vanilla GBL, our proposed UC-GBL has a couple of advantages: 1) The model enjoys a larger tolerance over prediction and label noise, and reduces the impact of noise which degrades the learning performance as the learned $\mathbf{\Sigma}$ and $\mathbf{W}$ can serve as attenuation; 2) eigenvalues and eigenvectors of $\mathbf{\Sigma}$ provide the localization uncertainty.

\comment{
Unlike \cite{he2019bounding, choi2019gaussian, luo2021rethinking, kendall2018multi}, we use covariance $\mathbf{\Sigma}$ to model the localization uncertainty for individual or multiple keypoints. Recall that the multivariate Gaussian distribution is $\mathcal{N}(\mathbf{x} | \mathbf{v}^*, \mathbf{\Sigma})$, 
where $\mathbf{x}, \mathbf{v^{*}} \in \mathbb{R}^{k}, k\geq2$, and $\mathbf{\Sigma} \in \mathbb{R}^{k \times k}$. If we regard $\mathbf{x}$ as the predicted offset for a keypoint or multiple keypoints and $\mathbf{v^{*}}$ as GT, we can construct the negative log-likelihood (NLL) loss as 
\begin{equation}\label{eq:nll-gaussian-loss}
  \begin{aligned}
      &\mathcal{L}_{\text{os-nll}} = -\mathbb{E}\log \mathcal{N}(\mathbf{x} | \mathbf{v}^*, \mathbf{\Sigma}) \\
      &\equiv \frac{1}{2}\mathbb{E}[(\mathbf{x} - \mathbf{v}^*)^{\intercal} \mathbf{\Sigma}^{-1} (\mathbf{x} - \mathbf{v}^*) + \log\det(\mathbf{\Sigma})].
  \end{aligned}
\end{equation}
However, by observing Eq.~\ref{eq:nll-gaussian-loss}, we notice $\mathcal{L}_{\text{os-nll}}$ requires to compute the inverse of covariance matrix $\mathbf{\Sigma}$, which is an obstacle in efficiency for back-propagation especially when order $k\geq 4$. To cercumvent this issue, instead of directly learning $\mathbf{\Sigma}$, we learn precision matrix $\mathbf{\Omega}$ as alternative, where $\mathbf{\Omega} = \mathbf{\Sigma}^{-1}$. Thus, $\mathcal{L}_{\text{os-nll}}$ can be rewritten as 
\begin{equation}\label{eq:nll-gaussian-loss2}
  \mathcal{L}_{\text{os-nll}} = \frac{1}{2}\mathbb{E}[(\mathbf{x} - \mathbf{v}^*)^{\intercal} \mathbf{\Omega} (\mathbf{x} - \mathbf{v}^*) - \log\det(\mathbf{\Omega})].
\end{equation}
As such, the $\mathcal{L}_{\text{os-nll}}$ can be easily computed as long as $\det(\mathbf{\Omega})$ to be positive. To guarantee this, we further let $\mathbf{\Omega} = \mathbf{Q}\mathbf{Q}^{\intercal}/d$, 
where $\mathbf{Q} \in \mathbb{R}^{k\times d}$ is the latent matrix from our covariance branch networks. In extreme case, an $\epsilon \rightarrow 0$ can be added to ensure $\det(\mathbf{\Omega}) > 0$. \emph{Consequently, we aim to learn the $\mathbf{Q}$ such that our pipeline could indirectly learn the covariance $\mathbf{\Sigma}$}. Firstly, we propose to learn the covariance for single keypoint by adding a covariance branch $\mathcal{G}_\text{v}$ to GBL (Fig.~\ref{fig:fskd-pipeline}), whose output is latent covariance field $\mathcal{G}_\text{v}(\mathbf{\Psi}_n) \in \mathbb{R}^{S \times S \times 2d}$ sharing same scale with grid probability map $\mathbf{p}$. Then the $\mathbf{Q} \in \mathbb{R}^{2 \times d}$ which encodes the covariance information for a keypoint can be extracted from grid $g$. Secondly, we also learn the relations for multiple keypoints within group (assume $m$ keypoints per group) by a multi-keypoint covariance branch $\mathcal{G}_\text{mkv}$, while its output is direct $\mathbf{Q}_\text{mkv} \in \mathbb{R}^{2m \times 2m}$. Therefore, we can obtain the corresponding $\mathbf{\Omega}_\text{mkv}$ and $\mathbf{\Sigma}_\text{mkv}$.

In addition to $\mathbf{\Sigma}$ which relfects the localization uncertainty over keypoints, following \cite{kendall2018multi, novotny2018self}, we also model the semantic uncertainty $\boldsymbol{\sigma}$ by learning SD map $\boldsymbol{\sigma}^{-1} \in \mathbb{R}^{H \times W \times 1}$ via SD head $\mathcal{D}$ (Fig.~\ref{fig:fskd-pipeline}). In fact, $\boldsymbol{\sigma}^{-1}$ is a single-channel map whose value ranges from 0 to 1. The higher SD, the more distinctive a image in visual perception. For each keypoint descriptor $\mathbf{\Psi}_n$, we can extract the corresponding SD as $w_n = (\boldsymbol{\sigma}_{\text{s}}^{-1}(\mathbf{u}_n) + \boldsymbol{\sigma}_{\text{q}}^{-1}(\mathbf{u}'_n))/2$, where $\boldsymbol{\sigma}_{\text{s}}^{-1}$ and $\boldsymbol{\sigma}_{\text{q}}^{-1}$ are the SD maps for support and query images; $\mathbf{u}_n$ and $\mathbf{u}'_n$ are the support and query keypoint positions. Let $\mathbf{W}=diag([w_1, w_1, \cdots, w_m, w_m])\in \mathbb{R}^{2m \times 2m}$ be a diagonal matrix whose diagonal entry is $w_n$. We incorperate $\mathbf{W}$ into loss function and thus Eq.~\ref{eq:nll-gaussian-loss2} can be extended as
\begin{equation}\label{eq:dev-uc-loss}
   \mathcal{L}_\text{uc}=\frac{1}{2}\mathbb{E}[(\mathbf{x} - \mathbf{v}^*)^{\intercal} (\mathbf{\Omega} + \beta\mathbf{W}) (\mathbf{x} - \mathbf{v}^*) - \log(\det(\mathbf{\Omega}\mathbf{W}^\beta))]
\end{equation}
where $\beta$ is the tradeoff and set to be $1$ for simplicity. In addition, we also use $w_n$ to reweight the cross-entropy grid classification loss $\mathcal{L}_{\text{cls-uc}}=-\mathbb{E}[\sqrt{w_n}\mathbb{I}_{[k=g^*]}\log(\mathbf{p})]$. Overall, with the $\mathcal{L}_\text{uc}$ and $\mathcal{L}_{\text{cls-uc}}$, we are able to establish the uncertainty-aided GBL, which is dubbed as UC-GBL.

Compared to vanilla GBL, the proposed UC-GBL has a few advantages: 1) The model has larger tolerance over prediction and label noise, and can reduce the impact of noise to degrade the learning performance as the learned $\mathbf{\Sigma}$ and $\mathbf{W}$ can serve as attenuation; 2) the covariance $\mathbf{\Sigma}$ can elegantly give the localization uncertainty by its eigenvalues and eigenvectors.
}

\subsubsection{Multi-scale UC-GBL and Uncertainty Fusion}
Increasing the scale $S$ will increase the precision of grids but also result in more grids. In order to reduce the risk of mislocalization, we employ multi-scale UC-GBL in our pipeline. Let the loss function at scale $S$ be $\mathcal{L}^{(S)} = \alpha_{1}\mathcal{L}_{\text{uc}} + \alpha_{2}\mathcal{L}_\text{cls-uc}$ (we set $\alpha_{1}=\alpha_{2} = 1$). Then the multi-scale localization loss $\mathcal{L}_\text{ms}$ is formulated as
\begin{equation}\label{eq:ms-loss}
  \mathcal{L}_\text{ms} = \frac{1}{N_S}\sum\nolimits_{i=1}^{N_S}  \mathcal{L}^{(S_i)},
\end{equation}
where $N_S$ is the number of scales used in FSKD. The unified keypoint prediction $\mathbf{u}$ is computed as
\begin{equation}
  \mathbf{u} = \frac{1}{N_S} \sum\nolimits_{i=1}^{N_S} \frac{l_{0}}{S_i}\Big(\mathbf{g}^{(S_i)}+0.5+0.5\mathbf{v}^{(S_i)}\Big),
\end{equation}
where $\mathbf{g}^{(S_i)}\!\in\!\{0,\cdots,S_i\!-\!1\}\!\times\!\{0,\cdots,S_i\!-\!1\}$ and $\mathbf{v}^{(S_i)}$ are the predicted 2D grid index and offset at scale $S_i$. The localization uncertainty $\mathbf{\Sigma}$ is obtained by
\begin{equation}
  \mathbf{\Sigma} = \frac{1}{4N_{S}}\sum\nolimits_{i=1}^{N_S} \Big(\frac{l_0}{S_i}\Big)^2 \mathbf{\Sigma}^{(S_i)},
\end{equation}
where $\mathbf{\Sigma}^{(S_i)}$ is the covariance at scale $S_i$ obtained by inverting the precision matrix $\mathbf{\Omega}^{(S_i)}$. A glance of unified estimated keypioints and uncertainty is shown in Fig. \ref{fig:fskd-pipeline}.

\comment{
Turning up the scale $S$ will increase the precision of grids but also bring in more grids. In order to reduce the risk of mislocalization, we employ multi-scale UC-GBL in our pipeline. Denoting the loss function at scale $S$ is $\mathcal{L}^{(S)}$, where $\mathcal{L}^{(S)} = \alpha_{1}\mathcal{L}_{\text{uc}} + \alpha_{2}\mathcal{L}_\text{cls-uc}$ and $\alpha_{1}$, $\alpha_{2} = 1$. Then the multi-scale localization loss $\mathcal{L}_\text{ms}$ is formulated as
\begin{equation}\label{eq:ms-loss}
  \mathcal{L}_\text{ms} = \frac{1}{N_S}\sum\nolimits_{i=1}^{N_S}  \mathcal{L}^{(S_i)},
\end{equation}
where $N_S$ is the number of scales used in FSKD. The unified keypoint prediction $\mathbf{u}$ is computed as 
\begin{equation}
  \mathbf{u} = \frac{1}{N_S} \sum\nolimits_{i=1}^{N_S} [g^{(S_i)}+0.5+0.5*\mathbf{v}^{(S_i)}]*l_{0} /S_i.
\end{equation}
$g^{(S_i)}$ and $\mathbf{v}^{(S_i)}$ is the predicted grid and offset at scale $S_i$. And, the localization uncertainty $\mathbf{\Sigma}$ can be obtained by
\begin{equation}
  \mathbf{\Sigma} = \frac{1}{4N_{S}}\sum\nolimits_{i=1}^{N_S} (\frac{l_0}{S_i})^2 \mathbf{\Sigma}^{(S_i)},
\end{equation}
where the $\mathbf{\Sigma}^{(S_i)}$ is the covariance at scale $S_i$ and can be achieved by simply inversing the precision matrix $\mathbf{\Omega}^{(S_i)}$. A glance of unified estimated keypioints and uncertainty is shown in Fig. \ref{fig:fskd-pipeline}.
}

\subsection{Learning with Auxiliary Keypoints}
\begin{figure}[!t]
    \centering
    \includegraphics[width=.98\linewidth]{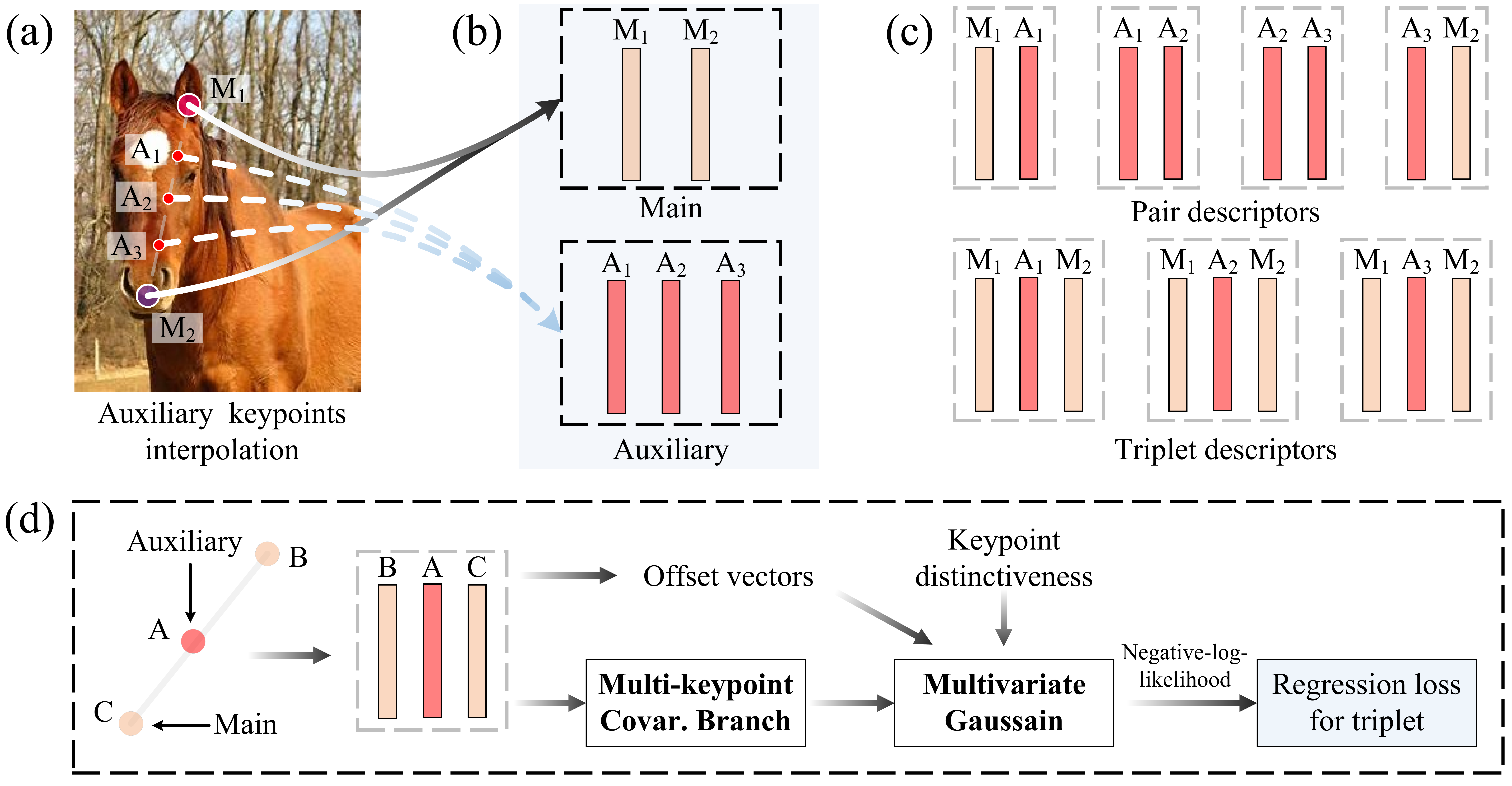}
    \caption{Keypoint grouping strategies and covariance modeling for multiple keypoints. (a) Interpolation with three nodes between body parts; (b) main and auxiliary descriptors; (c) grouping with pairs or triplets; (d) multi-keypoint covariance modeling.
    }
    \label{fig:multiple-descriptors}
    \vspace{-10pt}
\end{figure}
In addition to the main training keypoints provided by annotations, we adopt the auxiliary keypoints into learning, which are generated via interpolation $\mathcal{T}(t; [\mathbf{u}_1, \mathbf{u}_2])$ on a path whose end points are keypoints $[\mathbf{u}_1, \mathbf{u}_2]$, and $t\in(0,1)$ is the so-called interpolation node, 
as shown in Fig.~\ref{fig:multiple-descriptors}(a). We use an off-the-shelf saliency detector \cite{wu2019stacked} to prune auxiliary keypoints that do not lie on the foreground. Similarly, we build auxiliary descriptor $\tilde{\mathbf{\Psi}}_n$ (Fig.~\ref{fig:multiple-descriptors}(b)) and localization loss $\tilde{\mathcal{L}}_\text{ms}$ (Eq.~\ref{eq:ms-loss}) for auxiliary keypoints of query image. Even though auxiliary keypoints between support and query do not match well, they add visual diversity beyond appearances of training keypoints. We also group main and auxiliary keypoints into pairs or triplets (Fig.~\ref{fig:multiple-descriptors}(c)) and model the multi-keypoint covariance $\mathbf{\Sigma}_\text{mkv}$ via a branch $\mathcal{G}_\text{mkv}$ (Fig.~\ref{fig:multiple-descriptors}(d)). Thus, we propose a multi-keypoint offset regression loss $\mathcal{L}_\text{ms-mk}$.

\comment{
In addition to the main training keypoints provided by annotations, we adopt the auxiliary keypoints into learning, which are generated via interpolation $\mathcal{T}(t; [\mathbf{u}_1, \mathbf{u}_2])$ on a path whose end points are keypoints $[\mathbf{u}_1, \mathbf{u}_2]$, and $t\in(0,1)$ is the interpolation node, 
as shown in Fig.~\ref{fig:multiple-descriptors}(a). We use an off-the-shelf saliency detector \cite{wu2019stacked} to prune auxiliary keypoints that do not lie on the foreground. Similarly, we build auxiliary descriptor $\tilde{\mathbf{\Psi}}_n$ (Fig.~\ref{fig:multiple-descriptors}(b)) and localization loss $\tilde{\mathcal{L}}_\text{ms}$ (Eq.~\ref{eq:ms-loss}) for auxiliary keypoints on query image. Even though auxiliary keypoints between support and query does not match well, they add visual diversity beyond appearances of training keypoints. We also group main and auxiliary keypoints into triplets (or pairs) as shown in Fig.~\ref{fig:multiple-descriptors}(c) and model the multi-keypoint covariance $\mathbf{\Sigma}_\text{mkv}$ via a branch $\mathcal{G}_\text{mkv}$ (Fig.~\ref{fig:multiple-descriptors}(d)). Thus, we are able to construct multi-keypoint offset regression loss $\mathcal{L}_\text{ms-mk}$.
}
\comment{
In addition to the main training keypoints which are from annotations, we adopt the auxiliary keypoints into learning, which are generated via interpolation $\mathcal{T}(t; [\mathbf{u}_1, \mathbf{u}_2])$ on a path whose end points are body parts $[\mathbf{u}_1, \mathbf{u}_2]$, and $t$ is the interpolation node, as shown in Fig.~\ref{fig:multiple-descriptors}(a). Moreover, we use an off-the-shelf saliency detector \cite{wu2019stacked} to prune those invalid points falling out foreground. Similarly, we can build auxiliary descriptor $\tilde{\mathbf{\Psi}}_n$ (Fig.~\ref{fig:multiple-descriptors}(b)) and localization loss $\tilde{\mathcal{L}}_\text{ms}$ (Eq.~\ref{eq:ms-loss}) for auxiliary keypoints on query image. Though auxiliary keypoints between support and query do not match well, they help add diversity beyond main training keypoints. Besides, we also group main and auxiliary keypoints into triplets (or pairs) as shown in Fig.~\ref{fig:multiple-descriptors}(c) and then model the multi-keypoint covariance $\mathbf{\Sigma}_\text{mkv}$ via a branch $\mathcal{G}_\text{mkv}$ (Fig.~\ref{fig:multiple-descriptors}(d)). Thus, we are able to construct multi-keypoint offset regression loss $\mathcal{L}_\text{ms-mk}$.
}

\subsection{Objective Functions}
Our pipeline has three loss terms, which are the main keypoint loss $\mathcal{L}_\text{ms}$, auxiliary keypoint loss $\tilde{\mathcal{L}}_\text{ms}$, and multi-keypoint offset regression loss $\mathcal{L}_\text{ms-mk}$. $\mathcal{L}_\text{ms-mk}$ is used by default with the triplet grouping strategy. The overall loss ($\gamma_1=\gamma_2=\gamma_3=1$) is  
\begin{equation}
  \mathcal{L} = \gamma_1 \mathcal{L}_\text{ms} + \gamma_2 \tilde{\mathcal{L}}_\text{ms} + \gamma_3 \mathcal{L}_\text{ms-mk}.
\end{equation}

\comment{
In the proposed pipeline, three are three sources that contribute to the loss, which are main keypoint loss $\mathcal{L}_\text{ms}$, auxiliary keypoint loss $\tilde{\mathcal{L}}_\text{ms}$, and multi-keypoint offset regression loss $\mathcal{L}_\text{ms-mk}$. And $\mathcal{L}_\text{ms-mk}$ is in default using triplet grouping strategy. The overall objective function is  
\begin{equation}
  \mathcal{L} = \gamma_1 \mathcal{L}_\text{ms} + \gamma_2 \tilde{\mathcal{L}}_\text{ms} + \gamma_3 \mathcal{L}_\text{ms-mk}
\end{equation}
where $\gamma_1 = \gamma_2 = \gamma_3 = 0.5$ in our paper.
}


\section{Experiments}\label{sec:experiments}
\subsection{Experiment Setup}
\noindent\textbf{Datasets:}
1) \emph{Animal pose} \cite{cao2019cross} dataset has over 6,000 annotated instances with 5 species \emph{cat}, \emph{dog}, \emph{cow}, \emph{horse}, and \emph{sheep}. 17 body parts are used in our experiments; 2) \emph{CUB} \cite{WahCUB_200_2011} consists of 200 species with 15 keypoint annotations; 3) \emph{NABird} \cite{van2015building} has 555 categories, 11 keypoint annotations, and  48,562 images. For each dataset, we select the following keypoint types into the \emph{novel keypoint set}: i) \emph{two eyes} and \emph{four knees} for animal pose dataset; ii) \emph{forehead}, \emph{two eyes}, and \emph{two wings} for CUB; iii) \emph{two eyes} and \emph{two wings} for NABird. The remaining keypoints are used by the \emph{ base keypoint set}. 
See \S\ref{supp:sec:setup} of \textbf{Suppl. Material} (table with splits).

\comment{
There are three datasets employed in our experiments: 1) \emph{Animal dataset} \cite{cao2019cross} is extended from PASCAL VOC 2011 \cite{everingham2010pascal} and has five species, which are \emph{cat}, \emph{dog}, \emph{cow}, \emph{horse}, and \emph{sheep}. It has more than 6,000 instances annotated with body parts and 17 body parts are used in our experiments; 2) \emph{CUB dataset} \cite{WahCUB_200_2011} consists of 11,788 images from 200 species, and has annotated 15 keypoints; 3) \emph{NABird dataset} \cite{van2015building} owns 555 bird categories, 11 keypoints annotation, and in total 48,562 images, which is more challenging than CUB. To simply experiments, we split the following keypoint types into \emph{novel keypoint set} for each dataset: i) \emph{two eyes} and \emph{four knees} for animal pose dataset; ii) \emph{forehead}, \emph{two eyes}, and \emph{two wings} for CUB; iii) \emph{two eyes} and \emph{two wings} for NABird. The remaining keypoints are for \emph{base keypoint set}. For convenience, we also attach the concrete split table in supplementary materials for reference. 
}
\comment{
There are three datasets are employed in our experiments: 1) \emph{Animal pose dataset} \cite{cao2019cross} is extended from PASCAL VOC 2011 \cite{everingham2010pascal} and has five subsets of four-legged animals, which are \emph{cat}, \emph{dog}, \emph{cow}, \emph{horse}, and \emph{sheep}. It has in total more than 6,000 instances annotated with body parts, where 17 body parts are used in our experiments; 2) \emph{CUB dataset} \cite{WahCUB_200_2011} consists of 11,788 images from 200 species, 
and each instance has annotation of 15 body parts; 
3) \emph{NABird dataset} \cite{van2015building} is a more challenging dataset than CUB, and owns 555 bird categories and 48,562 images, where each instance is labelled with 11 body parts. In order to simply the experiments for few-shot keypoint detection, we split the keypoint category set of each dataset as shown in Table~\ref{tab:keypoints-split}. It should note that other splits are also applicable in our FSKD pipeline. 
\begin{table}[!tb]
  \centering
  \newcommand{\tabincell}[2]{\begin{tabular}{@{}#1@{}}#2\end{tabular}}
  \caption{Keypoint category splits for three datasets.}
  \label{tab:keypoints-split}
  \small
  \begin{tabular}{ccc}
    \toprule[1pt]
    Dataset & Base Keypoint Set & Novel Keypoint Set \\ \midrule
    Animal  & {\small\emph{\tabincell{l}{two ears, nose, four legs, \\four paws}}} & {\small\emph{\tabincell{l}{two eyes, four knees}}} \\ \midrule
    CUB     & {\small\emph{\tabincell{l}{beak, belly, back, breast, \\crown, two legs, nape, \\throat, tail}}} & {\small\emph{\tabincell{l}{forehead, two eyes, \\two wings}}} \\ \midrule
    NABird  & {\small\emph{\tabincell{l}{beak, belly, back, breast,\\ crown, nape, tail}}} & {\small\emph{\tabincell{l}{two eyes, two wings}}} \\ 
    \bottomrule
  \end{tabular}
\end{table}
}

\noindent\textbf{Metric:}
We use the percentage of correct keypoints (PCK) as evaluation metric \cite{yang2012articulated,novotny2018self}. The  distance of keypoint to GT should be less than $\tau\!\cdot\! \max ( w_{\text{bbx}}, h_{\text{bbx}} )$, where $w_{\text{bbx}}$ and $h_{\text{bbx}}$ are the edges of object bounding box. We set $\tau=0.1$.

\comment{
We use the percentage of correct keypoints (PCK) as evaluation metric \cite{yang2012articulated,novotny2018self}. A keypoint is regarded to be a correct prediction if its distance to the ground truth is less than $\tau \max \{w, h\}$, where $w$ and $h$ are the width and height of object bounding box. $\tau$ is the threshold and set to be $0.1$ throughout our experiments \cite{yang2012articulated,novotny2018self}.
}

\noindent\textbf{Compared Methods:}
We adapt for comparison the work of Novotny \etal \cite{novotny2018self} who developed the probabilistic introspection matching loss (\emph{ProbIntr}). Moreover, we build a \emph{baseline} using vanilla GBL at scale of $S=8$ (no auxiliary keypoints used, no uncertainty, \etc). For our FSKD approach, we introduce two variants for comparisons, which are \emph{FSKD (rand)} and \emph{FSKD (default)}. They share the architecture but differ by the types of interpolation path. The former, in each episode, randomly selects a number of paths formed by arbitrary two body parts for interpolation, while the latter interpolates auxiliary keypoints on limbs. 

\comment{
Since there are few existing methods directly targeting few-shot keypoint detection problem, we adapt the work proposed by Novotny \etal \cite{novotny2018self} for comparison, which is abbreviated as \emph{ProbIntr}. ProbIntr develops the so-called probabilistic introspection loss \cite{novotny2018self} to identify matching keypoints. 
Besides, we build a strong \emph{baseline} which uses vanilla GBL at scale of $S=8$ and without auxiliary keypoints training. For the proposed FSKD method, we introduce two variants for comparisons, which are \emph{FSKD (rand)} and \emph{FSKD (default)}. They share same architecture while differ on using which kind of interpolation path. The former, in each episode, randomly selects a number of paths formed by arbitrary two body parts for interpolation, while the latter interpolates auxiliary keypoints on limbs.  
}

\begin{table*}[!tb]
  \centering
  \newcommand{\tabincell}[2]{\begin{tabular}{@{}#1@{}}#2\end{tabular}}
  \caption{Results on 1-shot keypoint detection for unseen or seen species across three datasets. The PCK scores are reported.
  }
  \footnotesize
  \label{tab:kps-on-unseen-seen-species}
  \begin{tabular}{ccccccccccc}
      \toprule[1pt]
      \multirow{2}*{Species} & \multirow{2}*{Setting} & \multirow{2}*{Method}   & \multicolumn{6}{c}{Animal Pose Dataset}  & \multirow{2}*{CUB} & \multirow{2}*{NABird} \\\cmidrule(lr){4-9}
                                               &&             & Cat  & Dog  & Cow  & Horse & Sheep & Avg    &      &     \\ \midrule[1pt]
      \multirow{8}*{Unseen}&\multirow{3}*{Novel}& Baseline    &27.30 &24.40 &19.40 &18.25  &21.22  &22.11   &66.12 &39.14\\
                           &                    &ProbIntr     &28.54 &23.20 &19.55 &17.94  &17.03  &21.25   &68.07 &48.70\\
                           &                   &FSKD (rand)   &46.05 &40.66 &37.55 &38.09  &31.50  &38.77   &\textbf{77.90} &54.01\\
                           &                & FSKD (default)  &\textbf{52.36} &\textbf{47.94} &\textbf{44.07} &\textbf{42.77}  &\textbf{36.60}  &\textbf{44.75}   &77.89 &\textbf{56.04}\\\cmidrule(l){2-11}
                           &\multirow{3}*{Base} & Baseline    &51.08 &40.44 &45.27 &35.72  &43.03  &43.11   &81.16 &75.74\\
                           &                    &ProbIntr     &45.96 &42.49 &37.87 &40.53  &37.04  &40.78   &73.46 &70.56\\
                           &                   &FSKD (rand)   &\textbf{57.12} &51.12 &47.83 &49.71  &43.71  &49.90   &\textbf{87.94} &\textbf{87.84}\\
                           &                & FSKD (default)  &56.38 &\textbf{51.29} &\textbf{48.24} &\textbf{49.77}  &\textbf{43.95}  &\textbf{49.93}   &87.71 &86.99\\\midrule[1pt]
      \multirow{8}*{Seen}  &\multirow{3}*{Novel}& Baseline    &29.41 &24.43 &19.95 &19.59  &21.95  &23.07   &67.56 &43.52\\
                           &                    &ProbIntr     &26.09 &21.44 &19.71 &16.95  &17.83  &20.40   &64.13 &46.71\\
                           &                   &FSKD (rand)   &55.31 &44.08 &39.80 &41.52  &32.32  &42.61   &78.11 &56.33\\
                           &                & FSKD (default)  &\textbf{60.84} &\textbf{53.44} &\textbf{47.78} &\textbf{49.21}  &\textbf{38.47}  &\textbf{49.95}   &\textbf{78.17} &\textbf{58.35}\\\cmidrule(l){2-11}
                           &\multirow{3}*{Base} & Baseline    &62.30 &49.33 &51.33 &42.98  &44.18  &50.02   &84.02 &75.92\\
                           &                    &ProbIntr     &57.23 &48.58 &42.65 &48.70  &36.15  &46.66   &76.57 &70.47\\
                           &                   &FSKD (rand)   &67.55 &57.54 &\textbf{53.47} &57.40  &44.80  &56.15   &87.75 &87.88\\
                           &                & FSKD (default)  &\textbf{68.66} &\textbf{59.24} &52.70 &\textbf{58.53}  &\textbf{45.04}  &\textbf{56.83}   &\textbf{90.80} &\textbf{88.16}\\
      \bottomrule[1pt]
    \end{tabular}
\end{table*}
\begin{figure*}[!tb]
  \centering
  \includegraphics[width=.9\linewidth]{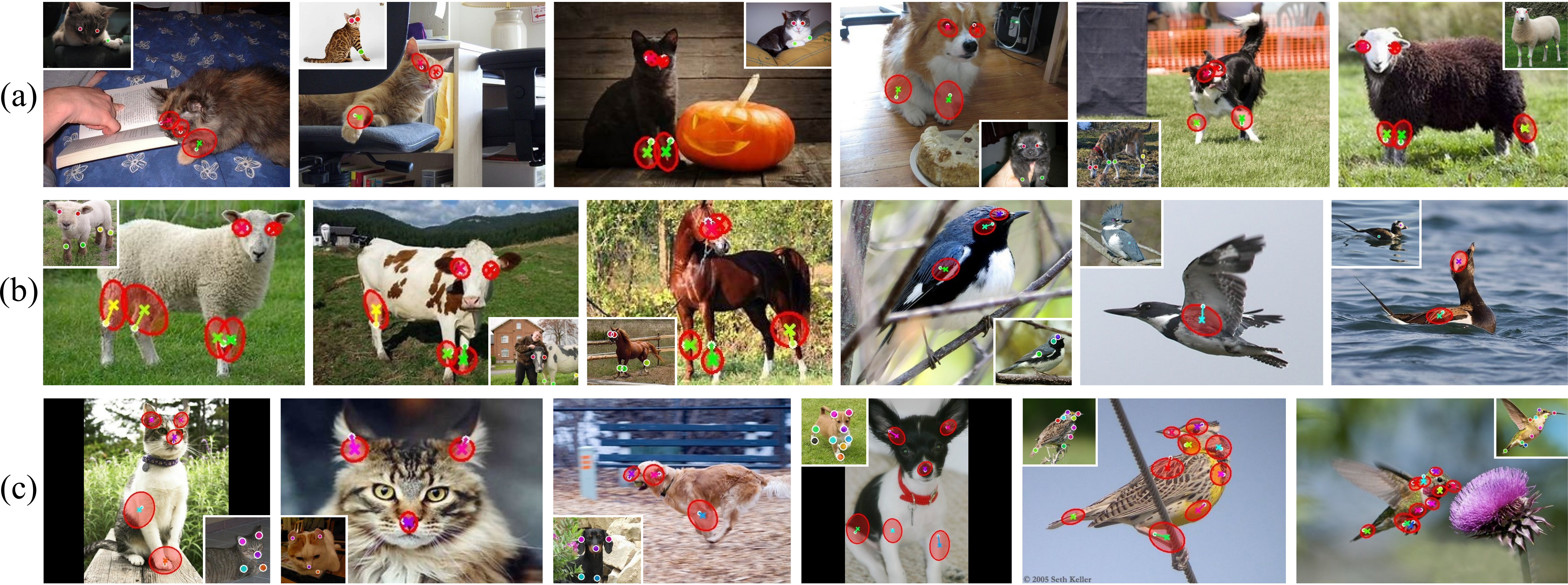}
  \caption{Examples of 1-shot keypoint detection for unseen species. The small image on the corner is the support while the big image is the query. The support keypoints and GT query keypoints are shown by circle dots. Our FSKD keypoint prediction is shown by tilted cross centered by an ellipse which represents the localization uncertainty with 99.7\% confidence. Not all support keypoints have the corresponding keypoints in query image as the GT may not exist. The line segment that connects from prediction to GT reveals the localization error. The rows (a)$\sim$(b) show the results on detecting novel keypoints while row (c) shows results for base keypoints.
}
  \label{fig:fskd-visual}
  \vspace{-2pt}
\end{figure*}

\noindent\textbf{Implementation Details:} 
Feature encoder $\mathcal{F}$ uses modified ResNet50 \cite{he2016deep}. 
Three UC-GBLs with scale $S=\{8, 12, 16\}$ are stacked to form the multi-scale UC-GBL. Random flips and rotations are used for data augmentation, and the image is resized and padded to $384 \times 384$. For auxiliary keypoints generation, the number of interpolation paths is set to  6 and the so-called interpolation nodes $t=\{0.25, 0.5, 0.75\}$. 
All the models are trained with 80k episodes in the animal pose dataset and 40k episodes in CUB and NABird. The results are reported over 1000 episodes in testing. 

\comment{
Our feature encoder $\mathcal{F}$ is modified from ResNet50 \cite{he2016deep}. to ensure the output feature map has downsize factor of $1/32$. We perform numerical transformation $f(x) = (x + \sqrt{x^2 + \epsilon})/2$ to guarantee semantic distinctiveness map $\sigma^{-1}$ to be positive. Three UC-GBLs at scale of $S=\{8, 12, 16\}$ are parallelly stacked to form multi-scale UC-GBL. The random flip and rotation is used for data augmentation and the image is resized and padded into $384 \times 384$. For auxiliary keypoints generation, we set the number of interpolation path to be 6 and the interpolation node $t$ to be $0.25, 0.5, 0.75$, respectively. All the models are trained with 80k episodes in animal pose dataset and 40k episodes in CUB and NABird. The results are reported over 1000 episodes in testing. We use Adam as optimizer and set the learning rate to be $1e-4$.
}
\comment{
Our feature encoder $\mathcal{F}$ is based on ResNet50 \cite{he2016deep} while dropping the rear layers behind Res4\_2 
to ensure the output feature map has a downsize factor of $1/32$ compared to input image. Since the weights pretrained on ImageNet \cite{deng2009imagenet} could provide stable low-level feature and help convergence, we fix the weights in first three convolutional (conv.) blocks. The SD head $\mathcal{D}$ is comprised of two conv. layers for downsizing and followed by a $1 \times 1$ conv. to convert the intermediate feature from Res3 into a single-channel semantic distinctiveness map $\sigma^{-1}$. Due to $\sigma^{-1}$ requires to be positive, we perform numerical transformation $f(x) = (x + \sqrt{x^2 + \epsilon})/2$ for the map $\sigma^{-1}$. The projector $\mathcal{P}$ starts and ends with separate conv. layer in purpose of adjusting channel, but the intermediate is a series of $3 \times 3$ conv. blocks which continuously reduce map size. 
In our UC-GBL $\mathcal{G}$, all branches are inplemented using MLP. And we parallelly stack three UC-GBL at scale of $S=8, 12, 16$ to perform multi-scale learning.

The random flip and rotation is used for data augmentation and the image is resized and padded into $384 \times 384$. For auxiliary keypoints generation, we set the number of interpolation path to be 6 and the interpolation node $t$ to be $0.25, 0.5, 0.75$, respectively. All the models are trained with 80k episodes in animal pose dataset and 40k episodes in CUB and NABird. The results are reported over 1000 episodes in testing. We use Adam as optimizer and set the learning rate to be $1e-4$. 
}

\begin{figure*}[!tb]
  \setlength{\abovecaptionskip}{0.05cm}
  \centering
  \begin{subfigure}[b]{0.19\linewidth}
    \centering
    \includegraphics[height=2.55cm]{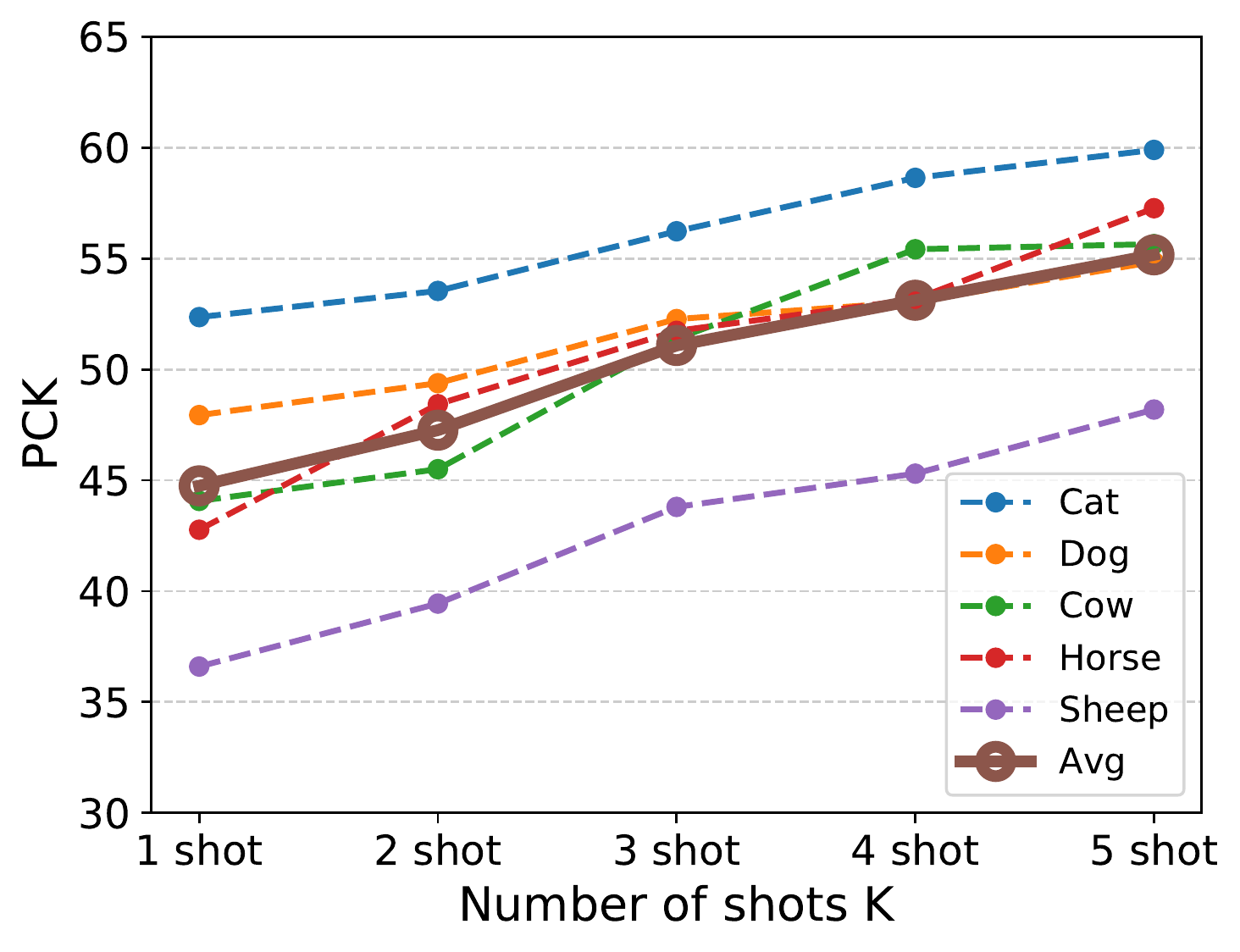}
    \subcaption{
    \label{fig:varying-shots-animals}}
  \end{subfigure}
  \begin{subfigure}[b]{0.19\linewidth}
    \centering
    \includegraphics[height=2.55cm]{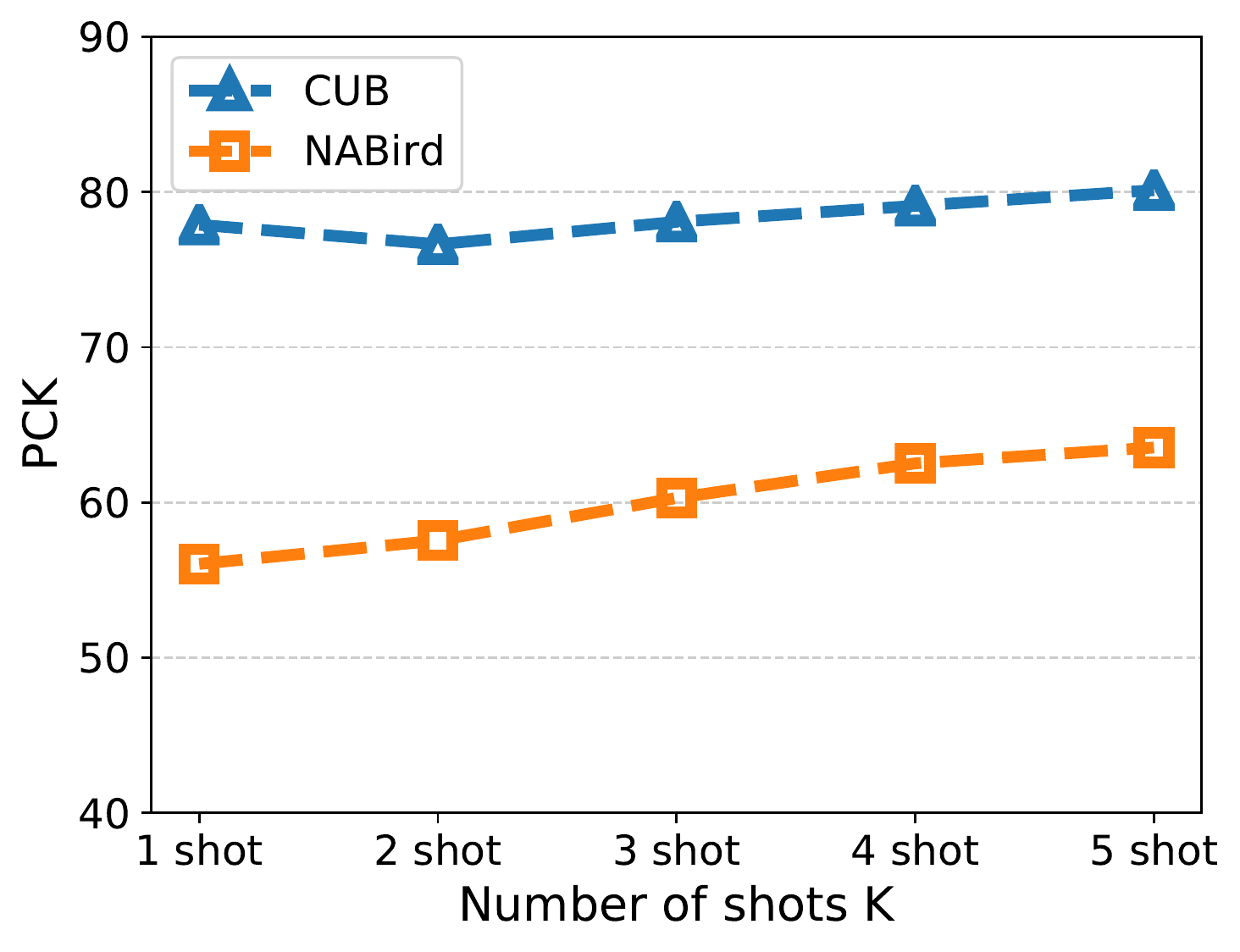}
    \subcaption{}
    \label{fig:varying-shots-birds}
  \end{subfigure}
  \begin{subfigure}[b]{0.19\linewidth}
    \centering
    \includegraphics[height=2.55cm]{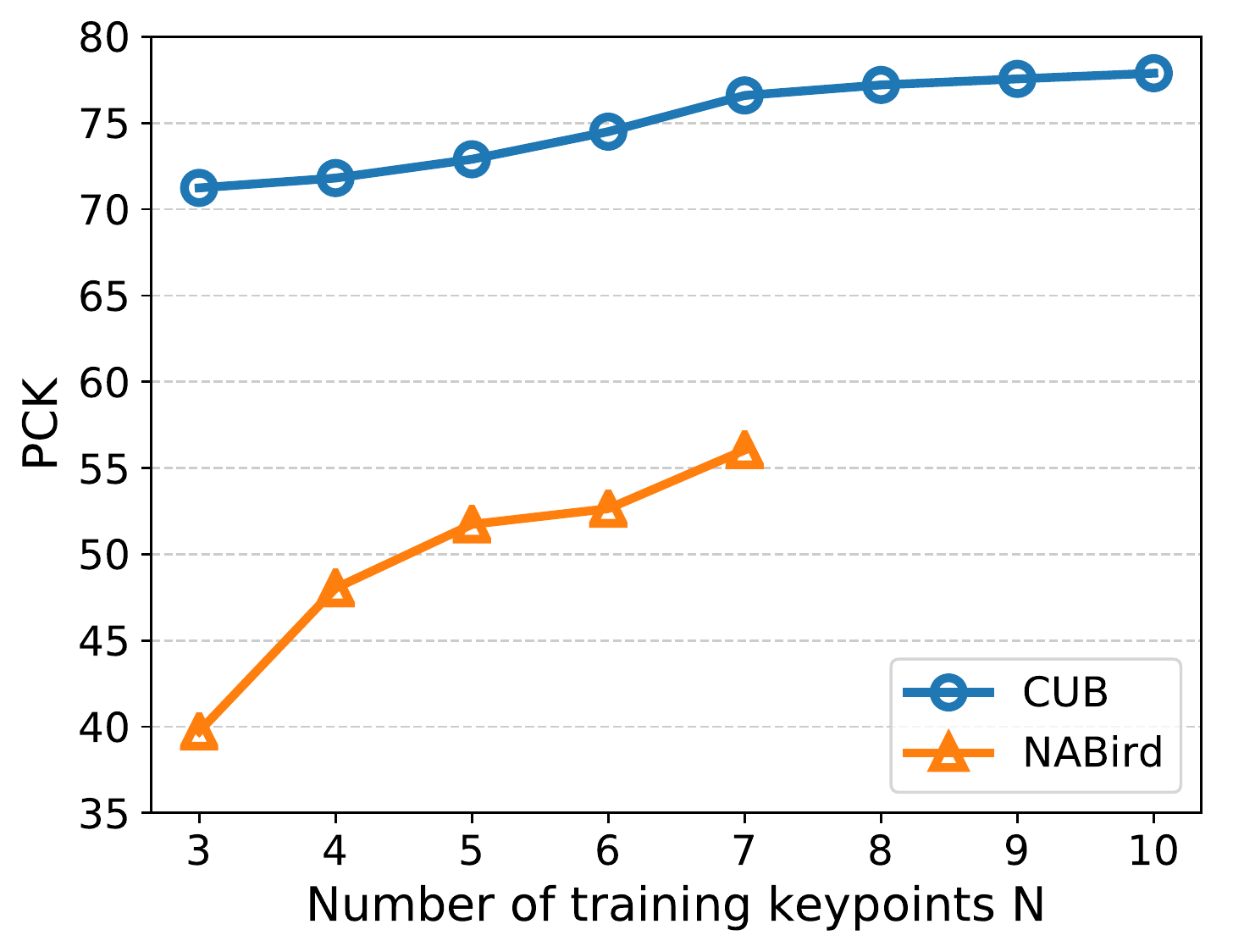}
    \subcaption{}
    \label{fig:varying-training-kps}
  \end{subfigure}
  \begin{subfigure}[b]{0.20\linewidth}
    \centering
    \includegraphics[height=2.55cm]{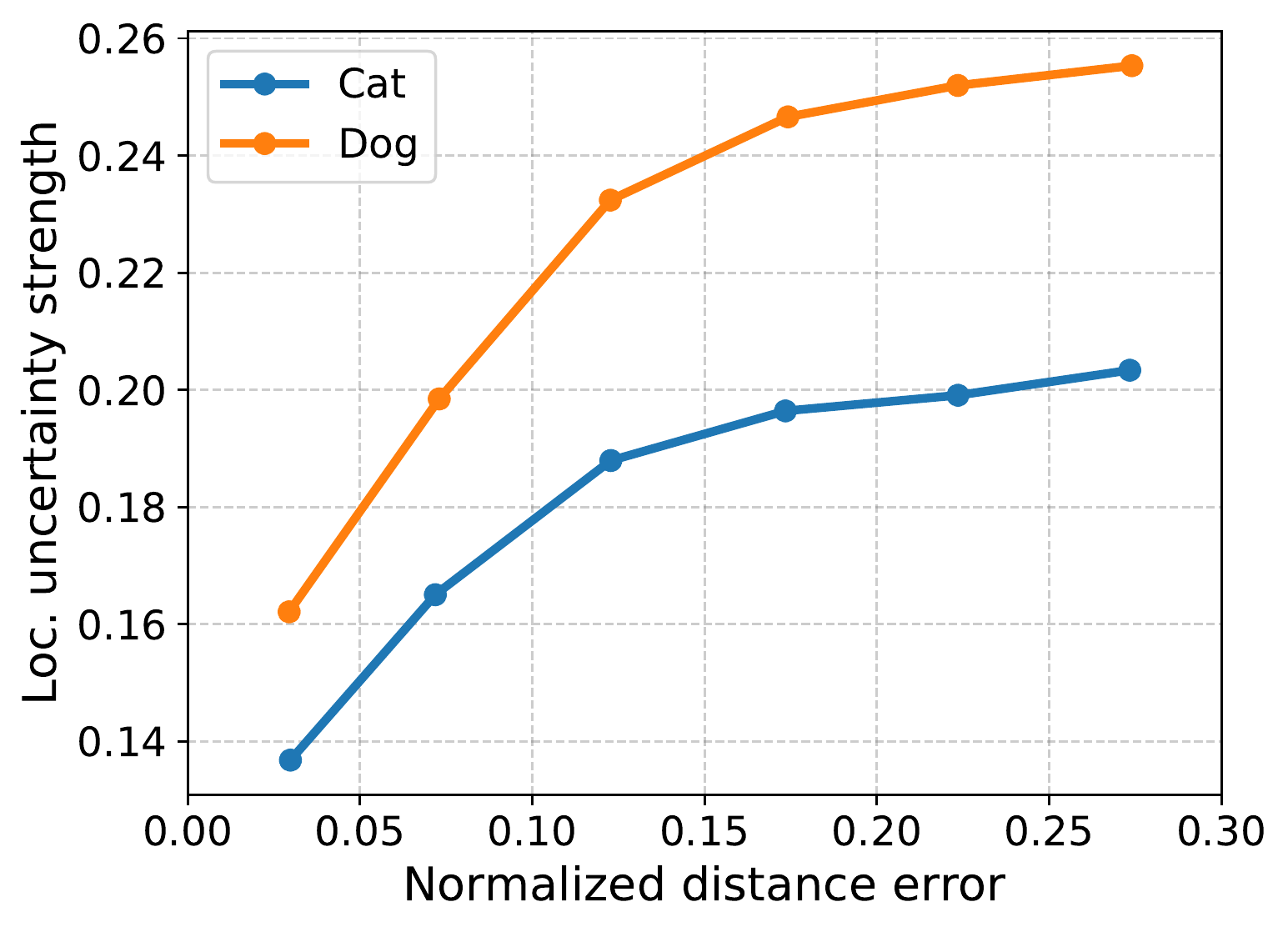}
    \subcaption{}
    \label{fig:uc-loc}
  \end{subfigure}
  \begin{subfigure}[b]{0.20\linewidth}
    \centering
    \includegraphics[height=2.55cm]{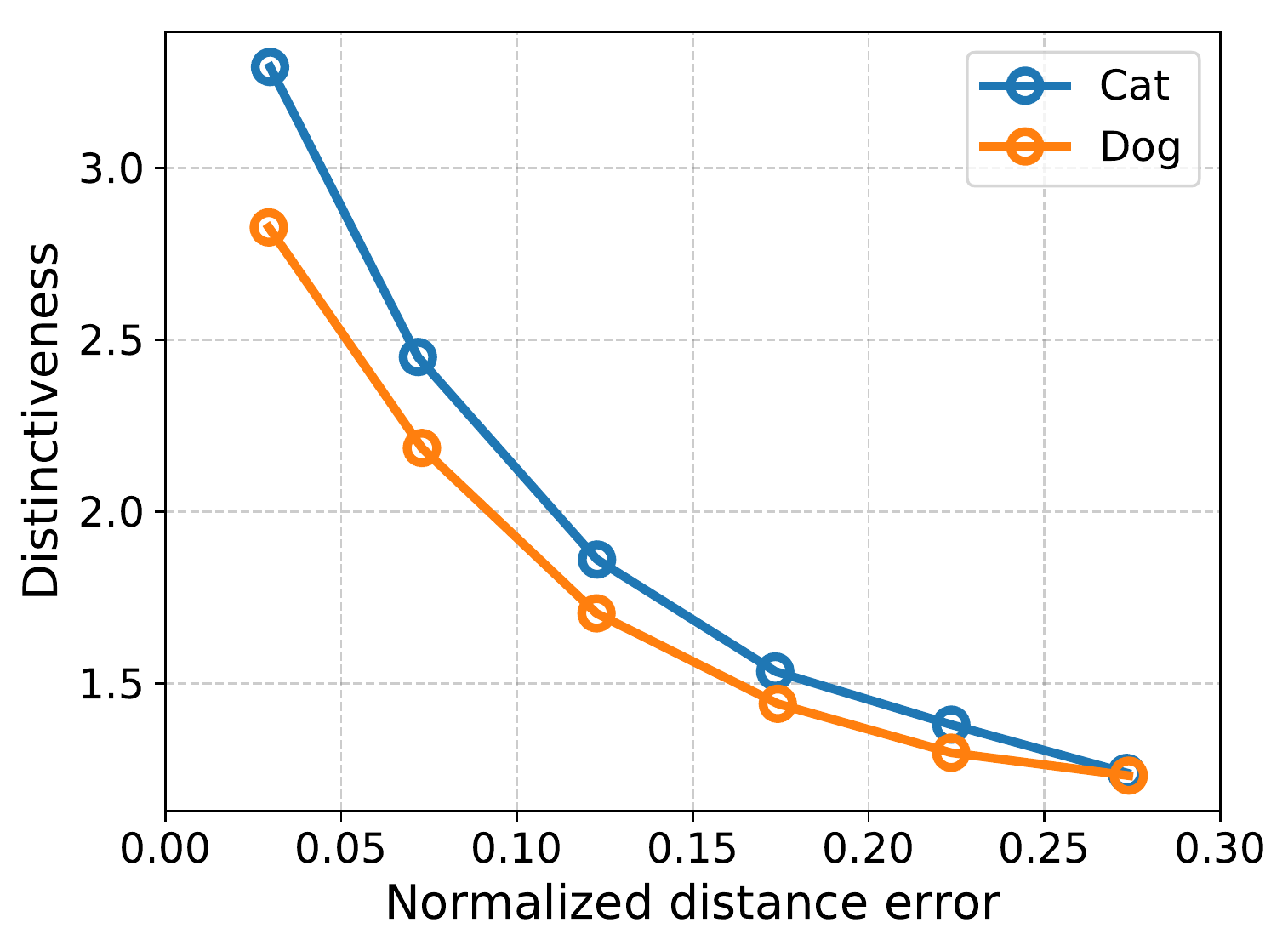}
    \subcaption{}
    \label{fig:uc-semantics}
  \end{subfigure}
  \caption{Study on shots, number of training keypoints, and uncertainty. (a)$\sim$(b) impact of shots $K$; (c) impact of number of training keypoints $N$; (d) and (e) show the localization uncertainty strength $J'$ and keypoint distinctiveness $w$ \vs the normalized distance error $d'$.
  }
  \label{fig:ablation-study}
  \vspace{-10pt}
\end{figure*}

\begin{table}[!tb]
  \centering
  \caption{FSKD (novel keypoints, unseen classes) on several backbones and 5 datasets. The score ($^*\!$27.75) is achieved by \emph{Baseline}.}
  \label{tab:backbones}
 \resizebox{0.48\textwidth}{!}{
\setlength{\tabcolsep}{3pt}
\fontsize{12}{12}\selectfont
  \renewcommand{\arraystretch}{0.5}
\begin{tabular}{cc|ccccc}
\toprule
     FSKD with     &  & Animal & CUB    & NABird  & DeepF.2  & AwA   \\ \midrule
$\!\!$\emph{ResNet50} &\cite{he2016deep}& 44.75  & 77.89  & 56.04   & 33.04    & $\!\!\!\!\!$\textbf{\color{red}64.76} ($^*\!$27.75)$\!\!\!$  \\
$\!\!$\emph{HRNet-W32} &\cite{sun2019deep} & 47.61& 78.24  & 56.89   & 33.67    & \textbf{\color{red}70.99} \\
$\!\!$\emph{HRNet-W48} &\cite{sun2019deep} & \textbf{48.81}& \textbf{79.45}  & \textbf{57.11}  & \textbf{34.29}  & \textbf{\color{red}72.20} \\ \bottomrule
\end{tabular}
 }
\end{table}

\begin{table}[!tb]
  \centering
  \newcommand{\tabincell}[2]{\begin{tabular}{@{}#1@{}}#2\end{tabular}}
  \caption{Ablation study on each component of FSKD. \emph{UC-GBL} is by default at scale $S=8$ and $^{*}$ means no use of $\mathcal{L}_{\text{ms-mk}}$. \emph{Aux} stands for adding auxiliary keypoints for training; \emph{MS UC-GBL} means multi-scale UC-GBL involved $S=\{8, 12, 16\}$. The results on the  Animal dataset are the average over five subproblems.
  }
  \small
  \label{tab:ablation-study}
  \resizebox{\hsize}{19mm}{  
  \begin{tabular}{lccc}
      \toprule[1pt]
      \multirow{1}*{One shot, PCK@$\tau=0.1$}          & Animal     & CUB    & NABird\\ \midrule[1pt]
        1: \emph{Baseline}                             &22.11       &66.12   &39.14  \\ \midrule
        2: \emph{Baseline+UC-GBL$^*$}                  &24.17       &68.29   &41.16  \\
        3: \emph{Baseline+UC-GBL$^*$+Aux.}             &41.70       &74.50   &51.62  \\ 
        4: \emph{Baseline+UC-GBL+Aux.}                 &42.60       &76.25   &54.27  \\
        5: \emph{Baseline+UC-GBL$^{(12)}$+Aux.}        &42.65       &76.90   &54.17  \\
        6: \emph{Baseline+UC-GBL$^{(16)}$+Aux.}        &42.61       &75.86   &54.15  \\ 
        7: \emph{Baseline+MS UC-GBL+Aux.}              &\textbf{44.75}&\textbf{77.89}&\textbf{56.04}  \\
      \bottomrule[1pt]
    \end{tabular}
   }
\vspace{-8pt}
\end{table}
\begin{figure}[!tb]
  \setlength{\abovecaptionskip}{0.1cm}
  \centering
  \includegraphics[width=\linewidth]{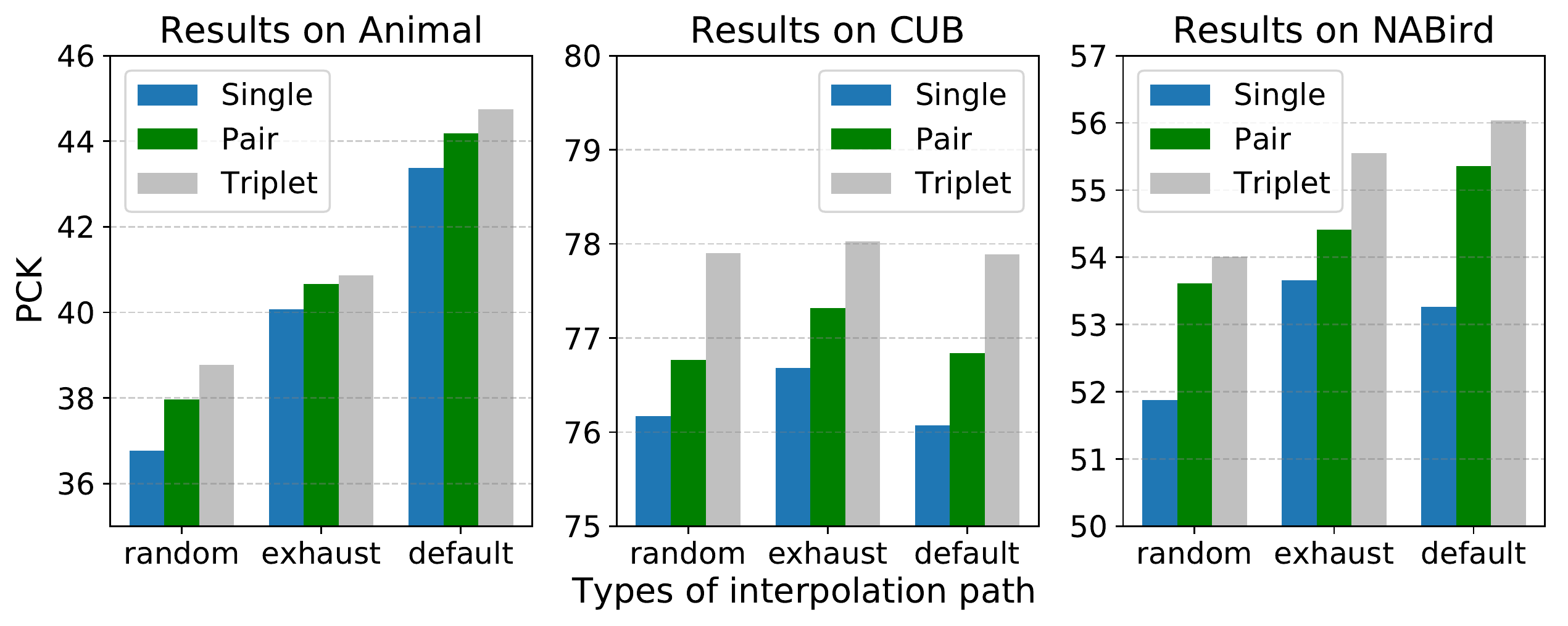}
  \caption{The impact of keypoint grouping strategies for various interpolation path strategies, where  \emph{single} means no grouping is used, and \emph{exhaust} means using the exhaustive path in interpolation.
  }
  \label{fig:interpolation-and-grouping}
  \vspace{-13pt}
\end{figure}

\subsection{Experiments on Few-shot Keypoint Detection}\label{sec:FSKD-experiments}

Firstly, we set experiments on detecting novel keypoints for unseen species (Fig.~\ref{fig:taxonomy}). For animal pose dataset, we alternately select one animal kind as unseen species for testing while the remaining four as seen species for training (five leave-one-out subproblems). For CUB, 100 species are for training, 50 for validation, and 50 for testing. In terms of NABird, the split is 333, 111, and 111 for training, validation and testing respectively. 
We use all-way 1-shot protocol on all tasks. In each episode, there is one support image and all base (or novel) keypoints are used as support keypoints. The same-species episode is used. Secondly, we explore detecting novel keypoints for seen species, where 70\% of images are for training and 30\% for testing (all datasets). Thirdly, for the above two settings, we also report testing results for base keypoints (using identical trained model). 

Table~\ref{tab:kps-on-unseen-seen-species} shows the results (95\% confidence intervals are below 1.2\%). We observe that: 1) The baseline is competitive compared to ProbIntr, which may be due to the success of feature modulator and vanilla GBL; 2) our FSKD variants significantly improve the scores in detecting novel keypoints for unseen species across three datasets, reaching 38.77\%, 77.90\%, 54.01\% for FSKD (rand) and 44.75\%, 77.89\%, 56.04\% for FSKD (default). 
Though FSKD (rand) uses noisy auxiliary keypoints, 
it performs very well in CUB and NABird and even surpasses FSKD (default) in CUB; 3) Results on base keypoints are higher than on novel keypoints due to smaller domain shift. Similar trend is observed for the seen species. 
The qualitative results by the FSKD (default) are shown in Fig.~\ref{fig:fskd-visual}. The localization uncertainty is represented by an ellipse whose major/minor axes are $(3\sqrt{\lambda_1}, 3\sqrt{\lambda_2})$, where $\lambda_i$ 
are eigenvalues of $\mathbf{\Sigma}$, and eigenvectors determine the orientation. Our FSKD  localizes keypoints well and estimates uncertainty which matches GT and correlates with the shape of body parts. 

\comment{
As illustrated in Fig.~\ref{fig:taxonomy}, firstly, we setup experiments on detecting novel keypoints for unseen species. For animal pose dataset, we alternately select one animal as unseen species for testing while the remaining four for training, which can result in five tasks. For CUB, 100 species are for training, 50 for validation, and 50 for testing. In terms of NABird, the split of species are 333, 111, and 111 for training, validation and testing respectively. \emph{It needs to stress that the base keypoints involved in training are disjoint with novel keypoints in testing and no external large-scale keypoint datasets are used for pretraining.} We perform all-way 1-shot for all tasks, which means that in each episode there is only one support image and all base (or novel) keypoints are as support keypoints. In addition, we use same-species episode in our experiments. Secondly, we explore the tasks of detecting novel keypoints for seen species, where 70\% of images of each species are for training and 30\% for testing. Thirdly, for the above two kinds of settings, we also use the identical trained model to report the results on base keypoints during testing. 

The comprehensive results are shown in Table~\ref{tab:kps-on-unseen-seen-species} and the 95\% confidence intervals are all below 1.2\%. We observe the findings as follows: 1) The baseline is competitive compared to ProbIntr, which may be due to the success of feature modulator and vanilla GBL; 2) the proposed FSKD variants could significantly improve the scores in detecting novel keypoints for unseen species across three datasets, reaching 38.77\%, 77.90\%, 54.01\% for rand version and 44.75\%, 77.89\%, 56.04\% for default version, which witnesses the power of FSKD with auxiliary keypoint and uncertainty learning. Though FSKD (rand) will introduce more noisy auxiliary keypoints when randomly selecting path for interpolation, it could perform very well in CUB and NABird and even surpass FSKD (default) in CUB; 3) When comparing the results between base and novel keypoints, it is obvious that the former is higher as smaller domain shift it has; 4) Similar trends could also oberve in seen species. However, detecting novel keypoints for seen species is still tough though it has gains in three datasets compared to unseen species.

The qualitative results by the FSKD (default) are visualized in Fig.~\ref{fig:fskd-visual}. The localization uncertainty is represented by an ellipse whose major/minor axis is proportional to the sqaure root of eigenvalues of $\mathbf{\Sigma}$, namely, $(3\sqrt{\lambda_1}, 3\sqrt{\lambda_2})$, and orientation is given by eigenvectors. As we can see, our FSKD is effective to localize keypoints and give uncertainty estimation which bounds GT well. Interestingly, the distribution of uncertainty shows relationship to the shape of body parts. 
}
\comment{
We will experiment our FSKD pipeline in all tasks illustrated in Fig.~\ref{fig:taxonomy} by leveraging the aforementioned datasets. Firstly, we setup experiments on detecting novel keypoints for unseen species, the task of which we are most interested in. Since there are merely five species in animal pose dataset, we alternately select one animal as unseen species for testing while the remaining four for training, which can result in five tasks. \emph{It needs to stress that the base keypoints involved in training are disjoint with novel keypoints in testing and no external large-scale keypoint datasets are used for pretraining.} For CUB dataset, we split it as 100 species for training, 50 for validation, and 50 for testing. In terms of NABird, 333 of species are used for training while both 111 are for validation and testing. We perform all-way 1-shot learning for all tasks, which means that in each episode all visible base keypoints are used for training and there is only one support image. Meanwhile, the experiments are run in same-species episode which refers that both support and query images are sampled from same species.

Secondly, we explore the tasks of detecting novel keypoints for seen species. To this end, each dataset will be splited as 70\% images for training, and the remaining 30\% images for testing. Both training set and testing set cover all species. As for the learning and episode setting, all are same to few-shot keypoint detection tasks in unseen species.

In addition to detecting novel keypoints, we also use the identical trained model to report the detection results on base keypoints during testing, separately. It is because FSKD framework is desired to detect any keypoints ideally as long as they are given in support image.

The comprehensive results are shown in Table~\ref{tab:kps-on-unseen-seen-species} and 95\% confidence intervals are all below 1.2\%. 1) By focusing on the first four rows, we observe that the baseline can achieve better PCKs than ProbIntr among most of tasks. This may be due to the baseline's feature modulator can successfully correlate features and the vanilla GBL is able to localize the keypoints from attentive features. Using the localization networks may be better than directly retrieving pixel-wise location and then matching as perfomed in ProbIntr \cite{novotny2018self}; 2) The proposed FSKD variants could significant improve the scores in detecting novel keypoints for unseen species across three datasets, reaching 38.77\%, 77.90\%, 54.01\% for rand version and 44.75\%, 77.89\%, 56.04\% for default version, which witness the power of auxiliary keypoint and multi-scale UC-GBL. It should note that though FSKD (rand) will introduce more noisy auxiliary keypoints when randomly selecting path for interpolation, it could perform very well in CUB and NABird and even surpass FSKD (default) in CUB; 
3) When comparing the detection results on unseen species between base and novel keypoints, we find that all methods consistently yield higher scores on base keypoint, indicating that the smaller domain shift it has. Moreover, our FSKD variants have quite close performance in these tasks; 4) As for the results on seen species, the similar trends can be observed for different methods. It is obvious that detecting novel keypoints is still a tough task though it obtains better results (which are 49.95\%, 78.17\%, 58.35\% in three datasets) compared to those in unseen species.

The qualitative results of 1-shot keypoint detection for unseen species by our FSKD approach are visualized in Fig.~\ref{fig:fskd-visual}. The localization uncertainty is represented by ellipse whose major/minor axis is proportional to the sqaure root of eigenvalues of $\mathbf{\Sigma}$, which is $3(\sqrt{\lambda_1}, \sqrt{\lambda_2})$. As we can see, our FSKD framework is effective to localize the keypoints and could give reliable estimation of localization uncertainty. Interestingly, the distribution of uncertainty also shows relationship to the shape of body parts.
}
%

{
\noindent\textbf{FSKD on DeepFashion2 \& AwA:} Table \ref{tab:backbones} shows results of  {FSKD} (default variant, 1-shot novel keypoint detection, unseen classes) \wrt 3 backbones/5 datasets, 
 including \emph{DeepFashion2} \cite{deepfashion2} (training on up-clothing categories/testing on lower-clothing categories) and the  diverse \emph{AwA Pose} \cite{banik2021novel} (novel keypoints types as in Animal test set, rest for training). 
Table \ref{tab:backbones} shows FSKD+HRNet-W48 yields 
 72.20\% on AwA Pose 
 ($\sim$23\% over 48.81\% of Animal dataset). 
}

\subsection{Ablation Study}\label{subsec:ablation-study}
Below we validate the effectiveness of each component using FSKD (default) under the novel keypoints detection. 


\noindent\textbf{Number of Shots:}
Typically, few-shot learning scores increase as the number of shots increases. 
Fig.~\ref{fig:ablation-study}\subref{fig:varying-shots-animals} and~\subref{fig:varying-shots-birds} show that PCK scores at 5-shot yield improvements of 10.42\% (average over five subproblems on Animal), 7.50\% in NABird, and 2.23\% in CUB, compared to 1-shot.

\comment{
An inherent nature of few-shot learning is that increasing the number of shots will bring in better scores since there are more observed samples to serve as references. As shown in Fig.~\ref{fig:ablation-study}\subref{fig:varying-shots-animals} and \subref{fig:varying-shots-birds}, PCK scores at 5-shot can yield improvements of 10.42\% in the average of animal dataset, 7.50\% over NABird, and 2.23\% over CUB, compared to 1-shot.
}
\comment{
An inherent nature of few-shot learning is that increasing the number of shots will bring in better scores since there are more observed samples to serve as references. To validate this, we increase the number of shots from 1 shot to 5 shot, and the resuls on three datasets are shown in Fig.~\ref{fig:varying-shots}. As we can see, our FSKD model is able to keep in pace with the increasing of shots. Specifically, the PCK scores at 5-shot can yield improvements of 10.42\% in the average of animal pose dataset, 7.50\% over NABird dataset, and 2.23\% over CUB dataset, compared to those results at 1-shot. Interestingly, it also shows increasing same number of shots may lead more gains for the model learned in harder dataset.  
}

\noindent\textbf{Number of Training Keypoints:}
We vary the number of training keypoints from base keypoint set and test the novel keypoints in CUB and NABird. Fig.~\ref{fig:ablation-study}\subref{fig:varying-training-kps} suggests that including more keypoints into training increases the keypoint diversity and helps FSKD generalize in novel keypoints.

\comment{
We use the varying number of keypoints from base keypoint set for training, and test the FSKD model on novel keypoints. For fair comparison, the number of testing keypoints are same for a group of experiments on a dataset. We conduct two groups of experiments on CUB and NABird respectively and the results are shown in Fig.~\ref{fig:varying-training-kps}. It could observe that it would be more preferable to include as more keypoints into training so as to increase the diversity and then help the model generalize better in novel keypoints. 
}

\noindent\textbf{Localization and Semantic Uncertainty:}
The statistical trend between the localization uncertainty and the distance error is shown in Fig. \ref{fig:ablation-study}\subref{fig:uc-loc}. 
We use $J=3(\sqrt{\lambda_1}+\sqrt{\lambda_2})$ to depict the `uncertainty strength' for a keypoint prediction, then normalize $J$ and distance error $d$ by bounding box as $(J'\!, d')\!=\!(J, d)/\max(w_{\text{bbx}}, h_{\text{bbx}})$. For keypoints, we divide $d'$ into intervals of size $0.05$ and calculate the average of $d'$ and average of corresponding $J'$ for each range. The plot shows $J'$ becomes larger as $d'$ increases, which validates that the learned uncertainty indicates the quality of predictions and can be used to suppress the noise in the loss function. Similarly, Fig. \ref{fig:ablation-study}\subref{fig:uc-semantics} shows the relation between the distinctiveness $w$ of keypoints and $d'$, that is, $w$ is lower when $d'$ is larger because 
a keypoint is harder to localize when it is less semantically distinctive. Including both localization and semantic uncertainty into our UC-GBL, the 2nd row of Table~\ref{tab:ablation-study} shows up to 3\% gain across three datasets.

\comment{
The statistical trend between localization uncertainty and distance error is shown in Fig. \ref{fig:ablation-study}\subref{fig:uc-loc}. We use the summation of uncertainty amplitudes, namely, $J=3(\sqrt{\lambda_1}+\sqrt{\lambda_2})$, to depict the overall uncertainty strength $J$ for a keypoint prediction, and its distance error $d$ is normalized by bounding boxes as $d'=d/\max\{w, h\}$. We divide $d'$ as increments of $0.05$ and calculate the average of $d'$ and the average of corresponding $J$ for each range. As we can see, $J$ will become larger as $d'$ increases, which validates that the learned uncertainty can indicate a prediction's quality and be used to attenuate the noisy loss. Similarly, we can draw the relation between keypoint prediction's distinctiveness $w$ versus $d'$ as shown in Fig. \ref{fig:ablation-study}\subref{fig:uc-semantics}, which says $w$ turns to be lower when $d'$ is larger. It is easy to understand as a keypoint is harder to localize when it is less distinctive in texture or semantics. When taking both localization and semantic uncertainty into our UC-GBL, as shown in 2nd row of Table~\ref{tab:ablation-study}, it brings in $2\%\sim 3\%$ gains across three datasets. 
}
\comment{
In fact, localization uncertainty gives the spatial distribution of confidence for keypoint predictions, and could attenuate the loss when a keypoint has large noise or error to groundtruth. To validate this, we first use the summation of uncertainty amplitudes, namely, $J=3(\sqrt{\lambda_1}+\sqrt{\lambda_2})$, to depict the overall uncertainty strength $J$ for a keypoint prediction, where $\lambda_i$ is the eigenvalue of $\mathbf{\Sigma}$. Then, the keypoint distance error $d$ is normalized by the longest edge of object bounding box as $d'=d/\max\{w, h\}$. We divide $d'$ as increments of $0.05$ and calculate the average of $d'$ and the average of corresponding $J$ for each range. The tendency between $J$ and $d'$ is shown in the left part of Fig.~\ref{fig:uc-analysis}. As we can see, the localization uncertainty strength $J$ will become larger as $d'$ increases within certain range, which indicates the learned uncertainty is able to serve as an effective indicator for the quality of a prediction and thus attenuate the loss from mislocalization. 

In addition, since our UC-GBL also incorperates the semantic uncertainty into loss function, we could also visualize statistical tendency of each keypoint prediction's distinctiveness $w$ versus $d'$. From the right part of Fig.~\ref{fig:uc-analysis}, we could see that the distinctiveness turns to be lower when a prediction poses larger error. It is easy to understand because a keypoint is harder to localize when the body part is less distinctive in texture or semantics. Some examples of semantic distinctiveness maps are shown in Fig.~\ref{fig:visual-distinctiveness}. Moreover, with both the localization and semantics uncertainty learning, the adoption of UC-GBL could bring in significant gains across three datasets, as shown in second row of Table~\ref{tab:ablation-study}.
}


\noindent\textbf{Analysis of Auxiliary Keypoints:}
Generating auxiliary keypoints (akin to self-supervision) boosts the visual diversity of training keypoints. The resulting noise is handled by our FSKD due to uncertainty modeling. After adding auxiliary keypoints, the scores improve dramatically (3rd row, Table~\ref{tab:ablation-study}). Moreover, when grouping auxiliary and main keypoints as triplets and modeling uncertainty for triplets, the performance improves further (4th row, Table~\ref{tab:ablation-study}). In addition, we also study the effects of different keypoint grouping strategies under various types of interpolation path. In Fig.~\ref{fig:interpolation-and-grouping}, we can see up to 3\% gain when using pairs or triplets.

\comment{
The usage of auxiliary keypoints can add the diversity for training samples and it is similar to self-supervision learning which mines the implicit information within data. Though the auxiliary keypoints have low-quality and poor matching between support and query as they are automatically generated, the uncertainty learned in our FSKD can handle this problem due to its advantage of noise suppression. Comparing second and third row of Table~\ref{tab:ablation-study}, we can see that adding auxiliary keypoints into learning dramatically boosts the scores, yielding the improvements of 17.53\%, 6.21\%, and 10.46\% in animal pose dataset, CUB, and NABird, respectively. Moreover, when grouping the auxiliary and main keypoints into triplets for learning, it could further improve the performance, as shown in fourth row in Table~\ref{tab:ablation-study}. 

Besides the aformentioned random and default types of path for interpolating auxiliary keypoints, we also explore the exhaust type which exhaustively uses all paths. The results are shown in Fig.~\ref{fig:interpolation-and-grouping}. Compared to the baseline in Table~\ref{tab:ablation-study}, the auxiliary keypoints from each type of path result in varying extent of benefits. Further, we could consistently observe the gain up to 3\% when using pair or triplet of keypoints (Fig.~\ref{fig:interpolation-and-grouping}), which reflects the usefulness of modeling covariance for multiple keypoints.
}

\noindent\textbf{Improvements on Multi-scale UC-GBL:}
Table~\ref{tab:ablation-study} shows that multi-scale UC-GBL outperforms  singe-scale models. Therefore, multi-scale learning limits mislocalization. 

\comment{
Multi-scale UC-GBL (MS UC-GBL) takes the predictions from each UC-GBL into overall consideration, thus it is able to alleviate the risk of mislocalization. Since our MS UC-GBL involves grid scales $S=8, 12, 16$, we perform ablation experiments for the FSKD configured with each single-scale UC-GBL and the results are shown in Table~\ref{tab:ablation-study}. It can see that the MS UC-GBL outperforms any single-scale model, which demonstrates the effectiveness of multi-scale learning. In addition, we visualize the output of each UC-GBL from MS UC-GBL, and an example is shown in Fig.~\ref{fig:multi-scale}. The keypoint predictions from multi-scale are indeed stabler (simply comparing the left eye at scale 12). With the increase of scale $S$, the uncertainty will shrink (shown in blue ellipses). However, our fused uncertainty (shown in red ellipse) can rightly combine those and gives a proper ensemble uncertainty estimation.
}

\vspace{-4pt}
\section{Downstream Tasks}
\vspace{-4pt}
\subsection{Few-shot Fine-grained Visual Recognition}
\vspace{-2pt}
\begin{table}[!tb]
  \centering
  \caption{Few-shot FGVR results. 
  \emph{AuxKps} means adding auxiliary keypoints to form augmented prototypes for classification.}
  \small
  \label{tab:FGVR}
  \begin{tabular}{llcccc}
    \toprule[1pt]
    \multirow{1}*{Datasets}& Models                 &            & 1-shot& 5-shot&all-shot\\ \midrule[1pt]
    \multirow{7}*{CUB}     & Proto &\cite{snell2017prototypical} &23.03  &38.05  &41.79\\
                           & Proto+BP &\cite{lin2015bilinear}    &18.43  &33.63  &38.34\\
                           & Proto+bbN &\cite{tang2020revisiting}&23.97  &40.22  &44.61\\
                           & Proto+PN &\cite{tang2020revisiting} &35.92  &58.66  &63.51\\
                           & Ours     &                          &\textbf{37.45}  &\textbf{61.22}  &\textbf{66.25}\\ 
                           & Ours+\emph{AuxKps}     &            &\textbf{38.04}  &\textbf{61.74}  &\textbf{66.37}\\ \midrule
    \multirow{2}*{NABird}  & Proto+PN &\cite{tang2020revisiting} &26.17  &50.55  &60.03\\ 
                           & Ours            &                   &\textbf{27.68}  &\textbf{51.81}  &\textbf{61.56}\\
    \bottomrule[1pt]
  \end{tabular}
  \vspace{-10pt}
\end{table}
Following \cite{tang2020revisiting}, we adopt pose normalization (PN) that uses the concatenation of body part features to capture the distinctive features across fine-grained classes (Fig. \ref{fig:fskd-downsteam-tasks}(a)). First, we modify  
our FSKD into a simple keypoint extractor by leveraging the universal keypoint prototype (UKP) computed by averaging the SKPs on additional 1000 episodes after training. In testing, UKPs guide FSKD to detect the keypoints, and thus FSKD \emph{no longer} needs the support input as a reference. We use FGVR model from \cite{tang2020revisiting}, based on ProtoNet (Proto) \cite{snell2017prototypical}, whereas our FSKD supplies keypoint predictions. Bilinear pooling (BP) \cite{lin2015bilinear} and bounding box normalization (bbN) \cite{tang2020revisiting} based methods are also compared. All  models are evaluated in all-way setting, 1-, 5- and all-shot results are reported. Table~\ref{tab:FGVR} shows our model  achieves the best scores which validates the 
quality of FSKD. Including the features from auxiliary keypoints further improves results by making the prototypes more discriminative. 

\comment{
Few-shot FGVR aims to classify query image given a set of support images from different while visually similar species. Following \cite{tang2020revisiting}, we adopt pose normalization (PN) that uses the concatenation of body part features to capture the distinctions across classes (Fig. \ref{fig:fskd-downsteam-tasks}(a)). We first modify our FSKD into a simple keypoint extractor by leveraging the universal keypoint prototype (UKP). Specifically, UKP is computed by averaging the SKPs in additional 1000 episodes after training. When testing, UKPs are used to guide FSKD to detect the corresponding keypoints, and thus FSKD \emph{no longer} needs supports as reference. We use FGVR architecture same to \cite{tang2020revisiting} which is based on prototypical networks (proto), while the keypoint predictions are from our FSKD. Besides, bilinear pooling (BP) \cite{lin2015bilinear} and bounding box normalization (bbN) based methods are also compared \cite{tang2020revisiting}. All the models are evaluated in all-way setting and the results of 1-shot, 5-shot, and all-shot are reported. As shown in Table~\ref{tab:FGVR}, our model consistently achieves best scores which validates the keypoint detection ability of proposed FSKD. Besides, when combining the features from auxiliary keypoints, it could further promote the scores since the prototypes may contain richer information to discriminate the fine-grained classes.
}
\comment{
Few-shot FGVR aims to classify query image given a set of support images from different while visually similar species. The key of task is to learn the subtle distinctions across classes. To this end, \cite{tang2020revisiting} revisits the pose normalization (PN) which uses the concatenation of body part features for few-shot FGVR, and shows that it could greatly improve the recognition ability. 
Thus, the keypoint detector is important for the correct extraction of body part features and to ensure the function of PN. In this section, we will show that, our FSKD can serve as simple while effective keypoint extractor and be used in Few-shot FGVR after minor modification.

Specifically, our FSKD will first train on all keypoints. Once the training has finished, an additional 1000 episodes will be randomly generated to compute the mean of each type of support keypoint prototype, which is called as universal keypoint prototype $\hat{\mathbf{\Phi}}_n$. When testing, the $\hat{\mathbf{\Phi}}_n$ will be used to guide FSKD to detect the correponding keypoint for each image, which means that \emph{our FSKD no longer needs supports as reference in testing} and can serve as a keypoint detector for specific keypoint types. 

Following \cite{tang2020revisiting}, we also use PN, ResNet18, and prototypical networks (proto) \cite{snell2017prototypical} for few-shot classification, while the keypoint predictions are from our FSKD. In addition, bilinear pooling (BP) \cite{lin2015bilinear} and bounding box normalization (bbN) based methods are also compared \cite{tang2020revisiting}. All the models are evaluated in all-way setting and the results of 1-shot, 5-shot, and all-shot are reported. As shown in Table~\ref{tab:FGVR}, the models equiped with our FSKD can achieve best scores consistently, which validates the keypoint detection ability of proposed FSKD. Further, when combining the features of auxiliary keypoints into prototypes, it could further promote the scores since the prototypes may contain richer information to discriminate the fine-grained classes.
}

\subsection{Semantic Alignment}

Semantic alignment (SA) is used in tasks such as recognition and graphics \cite{branson2014bird,shi2016robust}. We demonstrate that, under the imperfect keypoint predictions, the query image $\mathbf{I}_\text{q}$ can be warped into the rectified image $\mathbf{I}'_\text{q}$ which aligns well with the support image $\mathbf{I}_\text{s}$ (Fig. \ref{fig:fskd-downsteam-tasks}(b)).

\comment{
Semantic alignment usually rectifies an image towards the standard one in order to help recognition \cite{branson2014bird,shi2016robust}. The transformation utilizes thin-plate-spline (TPS) transformation \cite{bookstein1989principal} whose nonlinearity is able to transform various types of shape. In this section, we demonstrate that, under the imperfect keypoint predictions, the query image $\mathbf{I}_\text{q}$ can be warped into the rectified image $\mathbf{I}'_\text{q}$ which well semantically aligns to support image $\mathbf{I}_\text{s}$.
}

\noindent\textbf{Uncertainty-weighted Thin-plate-spline Warp:}
Unlike classic TPS warp \cite{bookstein1989principal, donato2003approximation}, 
the key idea of our uncertainty-weighted TPS warp is letting the unequal warping contributions of keypoints based on their uncertainty, as the well-matching correspondences should be encouraged to warp while uncertain ones not. Let $\mathbf{P}\!=\![\mathbf{p}_1, \cdots, \mathbf{p}_N]\!\in\!\mathbb{R}^{2\times N}$ be the support keypoints, $\mathbf{P}'\!=\![\mathbf{p}'_{1}, \cdots, \mathbf{p}'_{N}]\!\in\!\mathbb{R}^{2\times N}$ be the predicted query keypoints, and $\hat{\mathbf{P}}\!=\![\mathbf{1},\mathbf{P}^{\intercal}]^\intercal\!\in\! \mathbb{R}^{3\times N}$. 
Let the estimated uncertainty strength for each query keypoint be $J_i$, and $\mathbf{D}=diag([J_1, \cdots, J_N]) \in \mathbb{S}^{N}_{++}$ be a diagonal matrix. Then we obtain the transformation parameters\footnote[1]{Uncertainty-weighted TPS warp is derived in \S\ref{supp:sec:proof} of \textbf{Suppl. Material}.} $\mathbf{T} \in \mathbb{R}^{2 \times (N+3)}$ of uncertainty-weighted TPS warp as 
\begin{equation}\label{eq:uc-tps-warp}
  \mathbf{T} = \left(
    \left[
      \begin{array}{cc}
        \mathbf{R}+\lambda \mathbf{D}^{2} & \hat{\mathbf{P}}^\intercal \\
        \hat{\mathbf{P}}  & \mathbf{0}^{3\times 3}
      \end{array}
    \right]^{-1}
    \left[
      \begin{array}{c}
        \mathbf{P}'^{\intercal} \\
        \mathbf{0}^{3\times 2}
      \end{array}
    \right]
  \right)^{\intercal},
\end{equation}
Matrix $\mathbf{R} \in \mathbb{R}^{N\times N}$ 
has entries 
$r_{i,j}=d_{i,j}^{2}\log d_{i,j}^{2}$,   $d_{i,j}$ is the Euclidean distance between $\mathbf{p}_i$ and $\mathbf{p}_j$, and $\lambda\geq0$ is the warping penalty. 
By using $\mathbf{T}$, every pixel grid $\mathbf{q}'_{i}=[x_i, y_i]^\intercal$ in the rectified image $\mathbf{I}'_{\text{q}}$ has the mapped pixel in the query image $\mathbf{I}_{\text{q}}$ following the transformation $\mathbf{q}_{i}=\mathbf{T}\tilde{\mathbf{q}}'_{i}$, where $\tilde{\mathbf{q}}'_{i}=[r_{1,i}, \cdots, r_{N,i}, 1, x_i, y_i]^\intercal$ and $r_{n,i}=d_{n,i}^2\log d_{n,i}^2$, the $d_{n,i}$ is the Euclidean distance between the $n$-th support keypoint $\mathbf{p}_n$ and the pixel grid $\mathbf{q}'_{i}$. After image remapping $\mathbf{I}'_{\text{q}}[\mathbf{q}'_{i}]=\mathbf{I}_{\text{q}}[\mathbf{q}_{i}]$, we obtain the rectified image $\mathbf{I}'_{\text{q}}$. 

\comment{
Unlike classic TPS warp \cite{bookstein1989principal, donato2003approximation} which considers the equal importance for each pair of corresponding keypoints, the key idea of our uncertainty-weighted TPS warp is allowing the inequal warping degree among keypoints based on their uncertainty strength, as the good correspondences should be encouraged to warp while opposite for the bad ones. Denoting the support keypoints as $\mathbf{P}=[\mathbf{p}_1, \cdots, \mathbf{p}_N] \in \mathbb{R}^{2\times N}$ and the predicted query keypoints as $\mathbf{P}'=[\mathbf{p}'_{1}, \cdots, \mathbf{p}'_{N}] \in \mathbb{R}^{2\times N}$. Let the estimated uncertainty strength for each query keypoint as $J_i$, and $\mathbf{D}=diag([J_1, \cdots, J_N]) \in \mathbb{R}^{N \times N}$ is a diagonal matrix. Thus, we can obtain the transformation parameters\footnote[1]{The detailed proof of solution for uncertainty-weighted TPS warp can be found in \S\ref{supp:sec:proof} of \textbf{Suppl Material}.} $\mathbf{T} \in \mathbb{R}^{2 \times (N+3)}$ of uncertainty-weighted TPS warp as 
\begin{equation}\label{eq:uc-tps-warp}
\mathbf{T} = \left({\mathbf{L}}^{-1}
  \left[ 
  {\begin{array}{*{20}{c}}
  {{\mathbf{P}}{'^{\intercal}}}\\
  {\mathbf{0}^{3\times 2}}
  \end{array}}
  \right]
  \right)^{\intercal},
\end{equation}
where $\mathbf{L} \in \mathbb{R}^{(N+3) \times (N+3)}$ is determined by $\mathbf{P}$ and $\mathbf{D}$:
\begin{equation}
\mathbf{L} = \left[ 
  {\begin{array}{*{20}{c}}
  {\mathbf{R} + \lambda {\mathbf{D}^2}}&\mathbf{1}^{N \times 1}&{\mathbf{P}^{\intercal}}\\
  \mathbf{1}^{1 \times N}&\mathbf{0}&\mathbf{0}\\
  \mathbf{P}        &\mathbf{0}&\mathbf{0}
  \end{array}}
  \right].
\end{equation}
$\mathbf{R}$ is an $N\times N$ matrix 
and the entry on $i$-th row and $j$-th column is $r_{i,j}=d_{i,j}^{2}\log d_{i,j}^{2}$, where $d_{i,j}$ is the Euclidean distance between $\mathbf{p}_i$ and $\mathbf{p}_j$. $\lambda$ controls the penalty on warping. Using the solved $\mathbf{T}$, every pixel grid $\mathbf{q}'_{i}=[x_i, y_i]^\intercal$ on the rectified image $\mathbf{I}'_{\text{q}}$ can find the mapping pixel on the query image $\mathbf{I}_{\text{q}}$ by applying the transformation $\mathbf{q}_{i}=\mathbf{T}\tilde{\mathbf{q}}'_{i}$, where $\tilde{\mathbf{q}}'_{i}=[r_{1,i}, \cdots, r_{N,i}, 1, x_i, y_i]^\intercal$ and $r_{n,i}=d_{n,i}^2\log d_{n,i}$, the $d_{n,i}$ is the Euclidean distance between $n$-th support keypoint $\mathbf{p}_n$ and pixel grid $\mathbf{q}'_{i}$. After image remap $\mathbf{I}'_{\text{q}}[\mathbf{q}'_{i}]=\mathbf{I}_{\text{q}}[\mathbf{q}_{i}]$, we are able to acquire the rectified image $\mathbf{I}'_{\text{q}}$, which establishes the semantic alignment.
}
\comment{
Unlike the classic TPS warp \cite{bookstein1989principal, donato2003approximation} which considers the equal importance for each pair of corresponding keypoints, we propose a novel uncertainty-weighted TPS warp, the key idea of which is allowing the inequal warping degree among keypoints based on their uncertainty strength. In practice, the case usually happens since the good correspondences should be encouraged to warp while opposite for the bad ones. We denote the support keypoints by $\mathbf{P}=[\mathbf{p}_1, \cdots, \mathbf{p}_N] \in \mathbb{R}^{2\times N}$ and the predicted query keypoints by $\mathbf{P}'=[\mathbf{p}'_{1}, \cdots, \mathbf{p}'_{N}] \in \mathbb{R}^{2\times N}$, respectively. Let the estimated uncertainty strength for each query keypoint as $J_i$, and denote $\mathbf{D}=diag([J_1, \cdots, J_N]) \in \mathbb{R}^{N \times N}$ to be a diagonal matrix. Thus, we can obtain the transformation parameters\footnote[1]{The detailed proof of solution for uncertainty-weighted TPS warp can be found in supplementary materials.} $\mathbf{T} \in \mathbb{R}^{2 \times (N+3)}$ of uncertainty-weighted TPS warp as 
\begin{equation}\label{eq:uc-tps-warp}
\mathbf{T} = \left({\mathbf{L}}^{-1}
  \left[ 
  {\begin{array}{*{20}{c}}
  {{\mathbf{P}}{'^{\intercal}}}\\
  {\mathbf{0}^{3\times 2}}
  \end{array}}
  \right]
  \right)^{\intercal},
\end{equation}
where $\mathbf{L} \in \mathbb{R}^{(N+3) \times (N+3)}$ is a matrix determined by $\mathbf{P}$ and $\mathbf{D}$, which is
\begin{equation}
\mathbf{L} = \left[ 
  {\begin{array}{*{20}{c}}
  {\mathbf{R} + \lambda {\mathbf{D}^2}}&\mathbf{1}^{N \times 1}&{\mathbf{P}^{\intercal}}\\
  \mathbf{1}^{1 \times N}&\mathbf{0}&\mathbf{0}\\
  \mathbf{P}        &\mathbf{0}&\mathbf{0}
  \end{array}}
  \right].
\end{equation}
$\mathbf{R}$ is an $N\times N$ matrix 
and the entry on $i$-th row and $j$-th column is $r_{i,j}=d_{i,j}^{2}\log d_{i,j}^{2}$, where $d_{i,j}$ is the Euclidean distance between $\mathbf{p}_i$ and $\mathbf{p}_j$. $\lambda$ controls the penalty on warping. Using the solved $\mathbf{T}$, every pixel grid $\mathbf{q}'_{i}=[x_i, y_i]^\intercal$ on the rectified image $\mathbf{I}'_{\text{q}}$ can find the mapping pixel on the query image $\mathbf{I}_{\text{q}}$ by applying the transformation $\mathbf{q}_{i}=\mathbf{T}\tilde{\mathbf{q}}'_{i}$, where $\tilde{\mathbf{q}}'_{i}=[r_{1,i}, \cdots, r_{N,i}, 1, x_i, y_i]^\intercal$ and $r_{n,i}=d_{n,i}^2\log d_{n,i}$, the $d_{n,i}$ is the Euclidean distance between $n$-th support keypoint $\mathbf{p}_n$ and pixel grid $\mathbf{q}'_{i}$. After image remap $\mathbf{I}'_{\text{q}}[\mathbf{q}'_{i}]=\mathbf{I}_{\text{q}}[\mathbf{q}_{i}]$, we are able to acquire the rectified image $\mathbf{I}'_{\text{q}}$, which establishes the semantic alignment.
}

\begin{figure}[!tb]
  \setlength{\abovecaptionskip}{0.2cm}
  \centering
  \includegraphics[width=\linewidth]{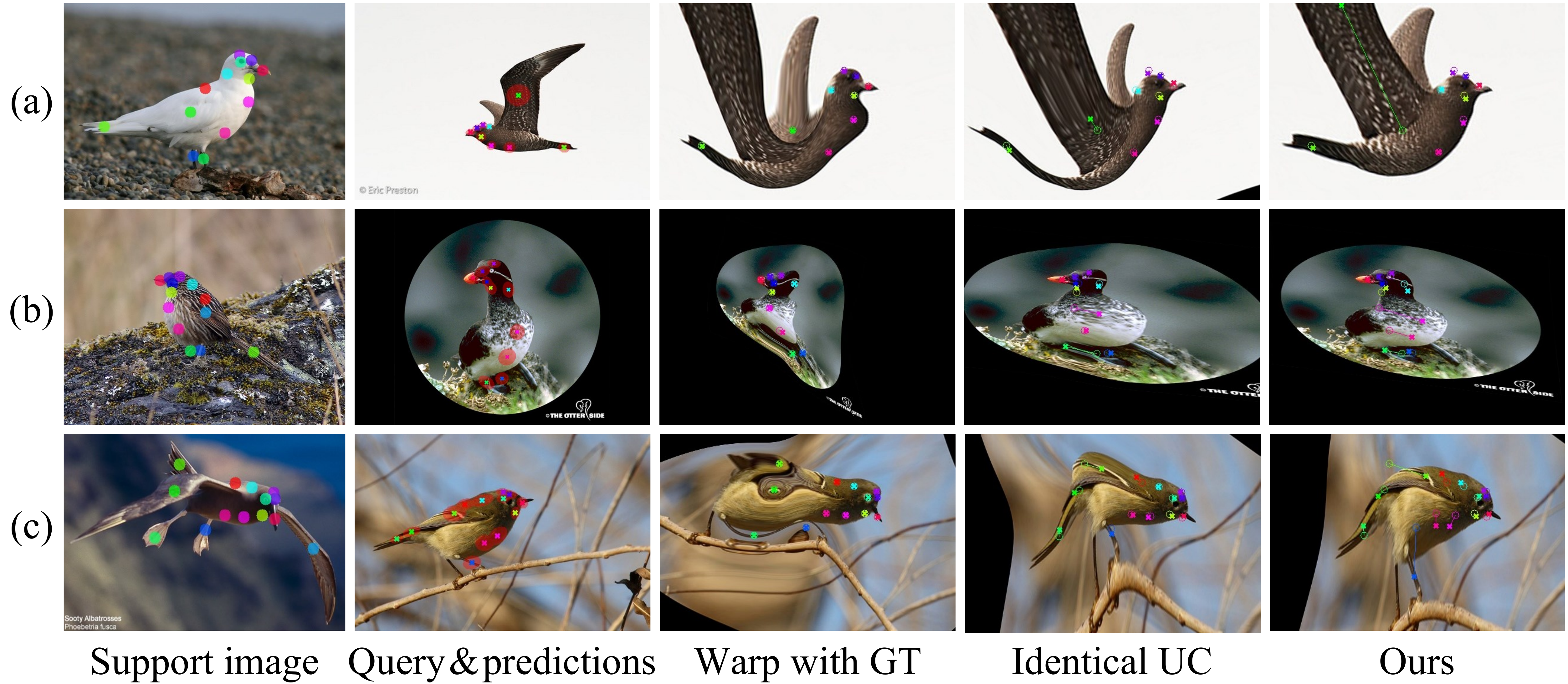}
  \caption{Comparison of semantic alignment  approaches. The first column shows the support keypoints \& image; the second column shows the query image with the predicted keypoints (marked by tilted crosses) and uncertainty (red shadow ellipses); the last three columns are the results achieved by \emph{Warp with GT} \cite{bookstein1989principal}, \emph{Identical UC}, and our uncertainty-weighted TPS warp. 
  }
  \label{fig:uc-tps-warp1}
  \vspace{-12pt}
\end{figure}

\vspace{0.05cm}
\noindent\textbf{Results:}
We perform SA for unseen species using 1-shot FSKD model trained on mix-species episodes. We set $\lambda\!=\!1$ and compare our approach with 1) \emph{Warp with GT} \cite{bookstein1989principal}, which uses GT query keypoints; and 2) \emph{Identical UC} using predicted keypoints with identical uncertainty $\mathbf{D}=diag([s,\cdots,s])$ where $s=20^2\log 20^2$ was chosen experimentally. 
Fig.~\ref{fig:uc-tps-warp1} shows that our approach penalizes the warp of uncertain keypoints, reduces the risks of unacceptable deformations, and produces a good alignment with support image.  Fig.~\ref{fig:uc-tps-warp1}(a) shows the detected wing (GT is right-wing) with large uncertainty due to the occlusion, which results in a small warp and  a large distance difference \wrt the corresponding support keypoint (Fig.~\ref{fig:uc-tps-warp1}(a), 5th column).

\comment{
We perform semantic alignment for unseen species in CUB/NABird datasets using the 1-shot FSKD model trained with mix-species episodes. Meanwhile, we compare our approach with 1) the traditional warp \cite{bookstein1989principal} which uses groundtruth (GT) query keypoints ($\lambda=0$); and 2) the warp with identical uncertainty using predicted keypoints by FSKD, where the $J=1$ and $\lambda=t^2\log t^2$, $t$ is set to be 20 due to its overall good performance. The qualitative results are shown in Fig.~\ref{fig:uc-tps-warp1}. Taking Fig.~\ref{fig:uc-tps-warp1}(b) as an example, the detected left-wing body part in query image has large uncertainty, and thus it results in the less warping on that part and maintains the replacement to the target support keypoint (Fig.~\ref{fig:uc-tps-warp1}(b), fifth column). As we can see, our approach can automatically penalize the bending degree for those uncertain keypoints, reduce the risks of disaster deformations, and align well to the supports, which demonstrates its better performance in semantic alignment. 
}    

\vspace{-2pt}
\section{Conclusion}\label{sec:conclusion}
\vspace{-1pt}
We have extended few-shot learning into the challenging task of keypoint detection by introducing a novel FSKD approach which learns the localization uncertainty of keypoints. FSKD is very flexible as it can detect keypoints of various types (seen \vs unseen) on various species (seen \vs unseen). Our simple uncertainty model deals with the keypoint noise and elegantly produces the uncertainty distribution of the likely position of GT keypoints. With the help of auxiliary keypoints, multi-keypoint covariance, and multi-scale localization, FSKD significantly boosts the detection performance. Moreover, FSKD can be successfully applied to a variety of downstream tasks such as FGVR and semantic alignment, where our novel uncertainty-weighted TPS warp leverages uncertainty. We hope our FSKD model will provide the starting point 
for the vision community and inspire the future research on few-shot keypoint detection.


\comment{
We have extended few-shot learning into the challenging task of keypoint detection by introducing a novel FSKD approach which learns the localization uncertainty of keypoints. FSKD is very flexible as it can detect keypoints of various types (seen \vs unseen) on various species (seen \vs unseen). Our simple and elegant uncertainty model  deals with the keypoint noise and produces the uncertainty distribution of the likely position of GT keypoints. With the help of auxiliary keypoints, multi-keypoint covariance, and multi-scale localization, FSKD significantly boosts the  detection performance. Moreover, FSKD can be successfully applied to the variety of downstream tasks such as FGVR and semantic alignment, where a novel uncertainty-weighted TPS warp benefits from uncertainty. Our FSKD is a starting point, a model on which others may build.
}




{\small
\bibliographystyle{ieee_fullname}
\bibliography{refs}
}

\newpage
\appendix
\title{Few-shot Keypoint Detection with Uncertainty Learning for Unseen Species (Supplementary Material)}

\author{Changsheng Lu$^{\dagger}$, Piotr Koniusz\textsuperscript{\textasteriskcentered}$^{,\S, \dagger}$\\
  $^{\dagger}$The Australian National University \quad 
   $^\S$Data61/CSIRO\\
{\tt\small ChangshengLuu@gmail.com, firstname.lastname@anu.edu.au}
}
\maketitle


{Our code will be released in the future: \url{https://github.com/AlanLuSun/Few-shot-keypoint-detection}.}

\section{Uncertainty-weighted TPS Warp}\label{supp:sec:proof}
\subsection{Proof of the Solution}
Let us denote two sets of fiducial points as $\mathbf{P}=[\mathbf{p}_1, \cdots, \mathbf{p}_N] \in \mathbb{R}^{2\times N}$ and $\mathbf{P}'=[\mathbf{p}'_{1}, \cdots, \mathbf{p}'_{N}] \in \mathbb{R}^{2\times N}$, where $\Pi_i =(\mathbf{p}_i, \mathbf{p}'_{i})$ is a pair of corresponding points. Let $\mathbf{W}=diag([w_1, \cdots, w_N]) \in \mathbb{S}_{++}^{N}$ be a diagonal matrix whose each diagonal entry $w_i$ is the confidence for $\Pi_i$. Then, our goal is to find a function mapping $f: \mathbf{p}_i\rightarrow \mathbf{p}'_i$ to achieve the minimal distance error between $\mathbf{P}$ and $\mathbf{P}'$ while ensuring the least deformation in rigidity. The objective of weighted TPS warp can be formulated as
\begin{equation}\label{supp-eq:tps-objective}
  \begin{aligned}
  &E(f, \mathbf{P}, \mathbf{P}') = \sum_{i=1}^{N} w_i^2\| \mathbf{p}'_i - f(\mathbf{p}_i) \|^2_2 + \\
  & \lambda \iint_{[x,y]^\intercal\in \mathbb{R}^2} \left(\frac{\partial^{2}f}{\partial x^2}\right)^{2} + \left(\frac{\partial^{2}f}{\partial y^2}\right)^{2} + 2\left(\frac{\partial^{2}f}{\partial x\partial y}\right)^{2}dxdy,
  \end{aligned}
\end{equation}
where the former term describes the weighted distance error, and the latter term (the definite integral) penalizes the so-called bending energy. Function $f$ can be constructed using the combination of affine transformation and a set of radial basis functions (RBF) as
\begin{equation}
  f(\mathbf{p}) = \mathbf{a}_{1} + \mathbf{a}_{2}x + \mathbf{a}_{3}y + \sum_{i=1}^{N}\mathbf{b}_i {\phi}(\|\mathbf{p}-\mathbf{p}_i\|_2),
\end{equation}
where $\mathbf{a}_i, \mathbf{b}_i\in \mathbb{R}^{2}$, $\mathbf{p}=[x, y]^\intercal \in \mathbb{R}^{2}$, and ${\phi}(\cdot)$ is a function with the radial basis. 
Following \cite{bookstein1989principal}, when choosing $\phi(d)=d^2\log d^2$, the bending energy in Eq.~\ref{supp-eq:tps-objective} can be minimized and the objective function could be reduced as
\begin{equation}
  E(f, \mathbf{P}, \mathbf{P}') =  \sum_{i=1}^{N} w_i^2\| \mathbf{p}'_i - \mathbf{A}\hat{\mathbf{p}}_i - \mathbf{B}\boldsymbol{\gamma}_i \|^2_2 + \lambda \text{tr}(\mathbf{B}\mathbf{R}\mathbf{B}^\intercal)
\end{equation}
where $\mathbf{A}=[\mathbf{a}_1, \mathbf{a}_2, \mathbf{a}_3] \in \mathbb{R}^{2\times 3}$, $\mathbf{B}=[\mathbf{b}_1, \cdots, \mathbf{b}_N] \in \mathbb{R}^{2\times N}$. $\hat{\mathbf{p}}_i = [1, \mathbf{p}_i^\intercal]^\intercal \in \mathbb{R}^{3}$ is the homogeneous coordinate. $\boldsymbol{\gamma}_i = [\gamma_{1,i}, \cdots, \gamma_{N, i}]^\intercal \in \mathbb{R}^{N}$ is a column vector containing the entry terms  $\gamma_{n,i}=d^2_{n,i}\log d^2_{n,i}$, and $d_{n,i}$ is the Euclidean distance between $\mathbf{p}_n$ and $\mathbf{p}_i$. $\mathbf{R} = [\boldsymbol{\gamma}_1, \cdots, \boldsymbol{\gamma}_N] \in \mathbb{S}_{++}^{N}$ is a symmetric positive definite matrix, \ie, $\mathbf{R}=\mathbf{R}^\intercal$ and $\mathbf{R}\succ 0$. By differentiating $E(f, \mathbf{P}, \mathbf{P}')$ w.r.t. $\mathbf{A}$ and $\mathbf{B}$, we have
\begin{equation}
\begin{aligned}
  &\frac{\partial E}{\partial \mathbf{A}} = \sum_{i=1}^{N} w_i^2 [-2\mathbf{p}'_{i}\hat{\mathbf{p}}_{i}^{\intercal} + 2\mathbf{B}\boldsymbol{\gamma}_i\hat{\mathbf{p}}_i^{\intercal} + 2\mathbf{A}\hat{\mathbf{p}}_{i}\hat{\mathbf{p}}_{i}^{\intercal}]  \\
  &\frac{\partial E}{\partial \mathbf{B}} = \sum_{i=1}^{N} w_i^2 [-2\mathbf{p}'_{i}\boldsymbol{\gamma}_{i}^{\intercal} + 2\mathbf{A}\hat{\mathbf{p}}_i\boldsymbol{\gamma}_i^{\intercal} + 2\mathbf{B}\boldsymbol{\gamma}_{i}\boldsymbol{\gamma}_{i}^\intercal] + 2\lambda \mathbf{B}\mathbf{R}
\end{aligned}.
\end{equation}
Let $\frac{\partial E}{\partial \mathbf{A}} = \mathbf{0}$ and $\frac{\partial E}{\partial \mathbf{B}} = \mathbf{0}$, we obtain the constraints as follows
\begin{align}
    (-\mathbf{P}'+\mathbf{B}\mathbf{R}+\mathbf{A}\hat{\mathbf{P}})\mathbf{W}^2\hat{\mathbf{P}}^\intercal &= \mathbf{0}  \label{supp-eq:tps-constraint1}\\
    \mathbf{A}\hat{\mathbf{P}} + \mathbf{B}(\mathbf{R}+ \lambda\mathbf{W}^{-2}) &= \mathbf{P}',  \label{supp-eq:tps-constraint2}
\end{align}
where $\hat{\mathbf{P}} = [\hat{\mathbf{p}}_1, \cdots, \hat{\mathbf{p}}_N]\in \mathbb{R}^{3\times N}$. By substituting $\mathbf{P}'$ in Eq.~\ref{supp-eq:tps-constraint1} with Eq.~\ref{supp-eq:tps-constraint2}, we have
\begin{equation}\label{supp-eq:tps-constraint1v2}
  \mathbf{B}\hat{\mathbf{P}}^\intercal = \mathbf{0}.
\end{equation}
Using Eq.~\ref{supp-eq:tps-constraint2} and Eq.~\ref{supp-eq:tps-constraint1v2}, we build the linear system as
\begin{equation}\label{supp-eq:uc-tps-linear}
  \left[
    \begin{array}{cc}
      \mathbf{R}+\lambda \mathbf{W}^{-2} & \hat{\mathbf{P}}^\intercal \\
      \hat{\mathbf{P}}  & \mathbf{0}^{3\times 3}
    \end{array}
  \right]
  \left[
    \begin{array}{c}
      \mathbf{B}^\intercal \\
      \mathbf{A}^\intercal
    \end{array}
  \right]=
  \left[
    \begin{array}{c}
      \mathbf{P}'^\intercal \\
      \mathbf{0}^{3\times 2}
    \end{array}
  \right].
\end{equation}
Solving the Eq.~\ref{supp-eq:uc-tps-linear}, the transformation parameters $\mathbf{T} = [\mathbf{B}, \mathbf{A}] \in \mathbb{R}^{2\times (N+3)}$ can be obtained.

If we have the uncertainty for each pair of points $\Pi_i$ as $J_i$, and denote $\mathbf{D}=diag([J_1, \cdots, J_N]) \in \mathbb{S}^{N}_{++}$ and $\mathbf{W}=\mathbf{D}^{-1}$, then the uncertainty-weighted TPS warp can be solved using Eq. \ref{supp-eq:uc-tps-linear}. The proof is completed.

\begin{figure}[!tb]
  \centering
  \includegraphics[width=\linewidth]{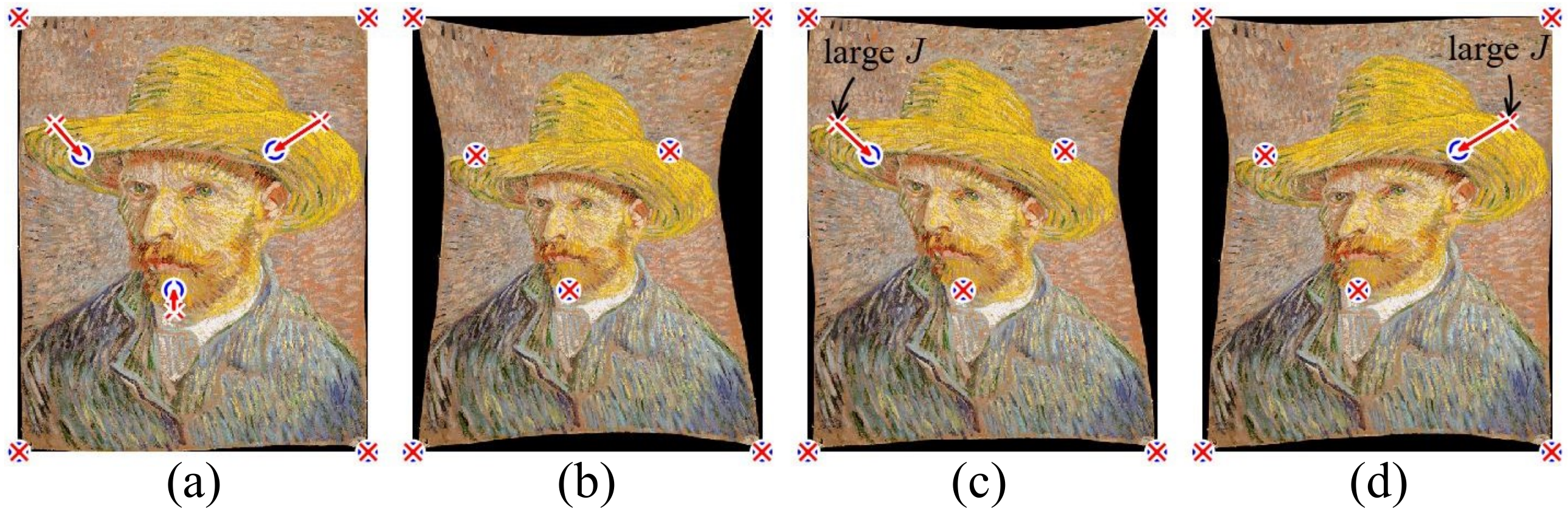}
  \caption{An example of using uncertainty-weighted TPS warp. (a) Original image that contains two sets of corresponding keypoints marked by red tilted crosses and blue circles; (b) perfect warping; (c) large uncertainty $J$ for the left keypoint on the hat; (d) large uncertainty $J$ for the right keypoint on the hat.}
  \label{fig:uc-tps-warp-example}
\end{figure}
\subsection{Toy Experiment}
To validate the effectiveness of uncertainty-weighted TPS warp, we warp the image given the corresponding keypoints as shown in Fig.~\ref{fig:uc-tps-warp-example}(a), where the tilted crosses are source keypoints and blue circles are target keypoints. After warping, the source keypoints will move to target keypoints. We manually set one keypoint correspondence to be with large uncertainty strength $J=100$ while other keypoints are with low uncertainty $J=1$. As shown in Fig.~\ref{fig:uc-tps-warp-example}(c) and Fig.~\ref{fig:uc-tps-warp-example}(d), the keypoint with the large uncertainty $J$ is less warped than other keypoints, which indicates that the proposed approach enjoys larger tolerance to the uncertain keypoints by focusing more on those confident keypoints.  

\section{Further Details for Experiment Setup}\label{supp:sec:setup}
\subsection{Compared Methods}
When modifying \emph{ProbIntr} \cite{novotny2018self} to adapt it to the few-shot keypoint detection task, we also build support keypoint prototypes (SKP) from extracted keypoint representations. The semantic distinctiveness (SD) is learnt  with the goal of constructing the probabilistic introspection matching loss $\mathcal{L}_{\text{m}}$ between SKP and individual query feature vectors. In addition to the matching loss from positive pair of keypoints, we also adopt negative pair of keypoints to perform hard negative mining. Moreover, we augment two views for each image in the episode and add the self-supervised loss $\mathcal{L}_{\text{ssl}}$ into the training step by randomly sampling 20 keypoints. The adopted feature encoder of \emph{ProbIntr} is ResNet50 which is identical to the encoder employed by our FSKD models. The whole network is trained by jointly optimizing $\mathcal{L}=\alpha \mathcal{L}_{\text{m}} + \mathcal{L}_{\text{ssl}}$, where $\alpha$ is set experimentally to $0.075$ (for the best performance of that baseline).

In our FSKD architecture, the output feature map of encoder has the size of $2048 \times 12 \times 12$ which indicates the downsize factor of $f=\frac{1}{32}$ compared to the image length. Since the model pretrained on ImageNet \cite{deng2009imagenet}  provides stable low-level features and helps convergence, we fix the weights of the first three convolutional (conv.) blocks of encoder. When using Gaussian pooling to extract keypoint representations, we set $\xi=14f=\frac{14}{32}$. The SD head consists of two conv. layers and a $1 \times 1$ conv. filter to convert the intermediate features into a single-channel SD map $\boldsymbol{\sigma}^{-1}$. We perform numerical transformation $f(x) = \frac{1}{2}(x + \sqrt{x^2 + \epsilon})$ to ensure SD map $\boldsymbol{\sigma}^{-1}>0$. The input and output of descriptor extractor contain dedicated conv. layers in order to manipulate their feature maps to desired sizes, 
whereas the intermediate layers contain a series of $3 \times 3$ conv. blocks which continuously reduce the feature map size. In our UC-GBL, all branches are implemented with MLP. We use the Adam optimizer and set the learning rate to $1e-4$.

\comment{
When modifying \emph{ProbIntr} \cite{novotny2018self} to adapt the few-shot keypoint detection task, we also build support keypoint prototypes (SKP) from extracted keypoint representations. The semantic distinctiveness (SD) is learned to construct the probabilistic introspection matching loss between SKP and individual query feature vectors. In addition to the matching loss from positive pair of keypoints, Novotny \etal \cite{novotny2018self} also adopt negative pair of keypoints to perform hard negative mining. In experiments, we find that using the negative pairs with top $k$ loss will degrade the few-shot keypoint detection performance, thus we alternatively sample random $k$ negative pairs, and $k=3$. The adopted feature encoder of \emph{ProbIntr} is ResNet50 which is same to our FSKD models.

In our FSKD architecture, the output feature map of encoder is with size of $2048 \times 12 \times 12$ which has downsize factor of $1/32$ compared to image length. Since the weights pretrained on ImageNet \cite{deng2009imagenet} could provide stable low-level features and help convergence, we fix the weights in first three convolutional (conv.) blocks of encoder. When using Gaussian pooling to extract keypoint representations, we set $\xi=14$. The SD head consists of two conv. layers and a $1 \times 1$ conv. filter to convert the intermediate features into a single-channel SD map $\boldsymbol{\sigma}^{-1}$. The descriptor extractor starts and ends with separate conv. layer in purpose of adjusting channel dimension, but the intermediate is a series of $3 \times 3$ conv. blocks which continuously reduce map size. In our UC-GBL, all branches are inplemented with MLP. We use Adam as optimizer and set the learning rate to be $1e-4$.
}


\subsection{Detailed Keypoint Splits}
We split the base keypoint set and the novel keypoint set of each dataset as detailed in   Table~\ref{tab:keypoints-split}. These splits are used in our experiments. We notice that other split choices could also be used in our FSKD pipeline.
\begin{table}[!tb]
  \centering
  \newcommand{\tabincell}[2]{\begin{tabular}{@{}#1@{}}#2\end{tabular}}
  \caption{Keypoint splits used in our experiments for three datasets.}
  \label{tab:keypoints-split}
  \small
  \begin{tabular}{ccc}
    \toprule[1pt]
    Dataset & Base Keypoint Set & Novel Keypoint Set \\ \midrule
    Animal  & {\small\emph{\tabincell{l}{two ears, nose, four legs, \\four paws}}} & {\small\emph{\tabincell{l}{two eyes, four knees}}} \\ \midrule
    CUB     & {\small\emph{\tabincell{l}{beak, belly, back, breast, \\crown, two legs, nape, \\throat, tail}}} & {\small\emph{\tabincell{l}{forehead, two eyes, \\two wings}}} \\ \midrule
    NABird  & {\small\emph{\tabincell{l}{beak, belly, back, breast,\\ crown, nape, tail}}} & {\small\emph{\tabincell{l}{two eyes, two wings}}} \\
    \bottomrule
  \end{tabular}
\end{table}

\section{Additional FSKD Results}

\begin{table}[!tb]
  \centering
  \caption{Additional comparison results on 1-shot novel keypoint detection.}
  \label{tab:more-comparison-results}
{
\hspace{-0.5cm}
  \resizebox{0.5\textwidth}{!}{
  \setlength{\tabcolsep}{3pt}
  \fontsize{12}{12}\selectfont
  \begin{tabular}{cccccccccc}
      \toprule
      \multirow{2}*{Method} & \kern-2.2em & \multicolumn{6}{c}{Animal Pose Dataset}  & \multirow{2}*{CUB} & \multirow{2}*{NABird} \\\cmidrule(lr){2-7}
                            &\kern-2.2em  & Cat  & Dog  & Cow  & Horse & Sheep & Avg    &      &     \\\midrule
                  Baseline  & \kern-2.2em &27.30 &24.40 &19.40 &18.25  &21.22  &22.11   &66.12 &39.14\\
                ProbIntr &\kern-2.2em\cite{novotny2018self}&28.54 &23.20 &19.55 &17.94  &17.03  &21.25   &68.07 &48.70\\
                TFA &\kern-2.2em\citelatex{suppl_wang2020frustratingly}&19.40 &20.00 &20.85 &17.99  &19.54  &19.56   &50.12 &30.16\\
                ProtoNet &\kern-2.2em\cite{snell2017prototypical}&19.68 &16.18 &14.39 &12.05  &15.06  &15.47   &51.32 &36.65\\
                RelationNet &\kern-2.2em\cite{sung2018learning}&22.15 &17.19 &15.47 &13.58  &16.55  &16.99   &56.59 &34.02\\
                WG (w/o Att.) &\kern-2.2em\citelatex{suppl_gidaris2018dynamic}&21.86 &17.11 &16.19 &16.34  &16.13  &17.53   &52.66 &33.31\\
                WG &\kern-2.2em\citelatex{suppl_gidaris2018dynamic}&22.47 &19.39 &16.82 &16.40  &16.94  &18.40   &54.75 &34.19\\
          FSKD (rand) (Ours)  & \kern-2.2em  &46.05 &40.66 &37.55 &38.09  &31.50  &38.77   &77.90 &54.01\\
          FSKD (default) (Ours) & \kern-2.2em&52.36 &47.94 &44.07 &42.77  &36.60  &44.75   &77.89 &56.04\\ \bottomrule
    \end{tabular}
    }
}
\end{table}

Popular few-shot learning (FSL) methods \eg, \emph{ProtoNet} \cite{snell2017prototypical}, \emph{RelationNet} \cite{sung2018learning}, two versions of \emph{WG} (with or without attention) \citelatex{suppl_gidaris2018dynamic}, and Two-stage Finetuning Approach (\emph{TFA}) \citelatex{suppl_wang2020frustratingly} are adapted to perform the FSKD task. 
All methods use the ResNet50 backbone and are evaluated under the setting of 1-shot novel keypoint detection (\textbf{Sec.} \ref{sec:FSKD-experiments}). Table \ref{tab:more-comparison-results} shows that the adapted state-of-the-art FSL approaches struggle to learn from the limited number of base keypoints. In constrast, thanks to the novel FSKD-specific designs such as single/multi-keypoint uncertainty modeling, auxiliary keypoints learning, and multi-scale UC-GBL, our FSKD variants achieve the best performance and outperform the above baselines by a large margin.


\begin{figure*}[!tb]
  \centering
  \includegraphics[width=.94\linewidth]{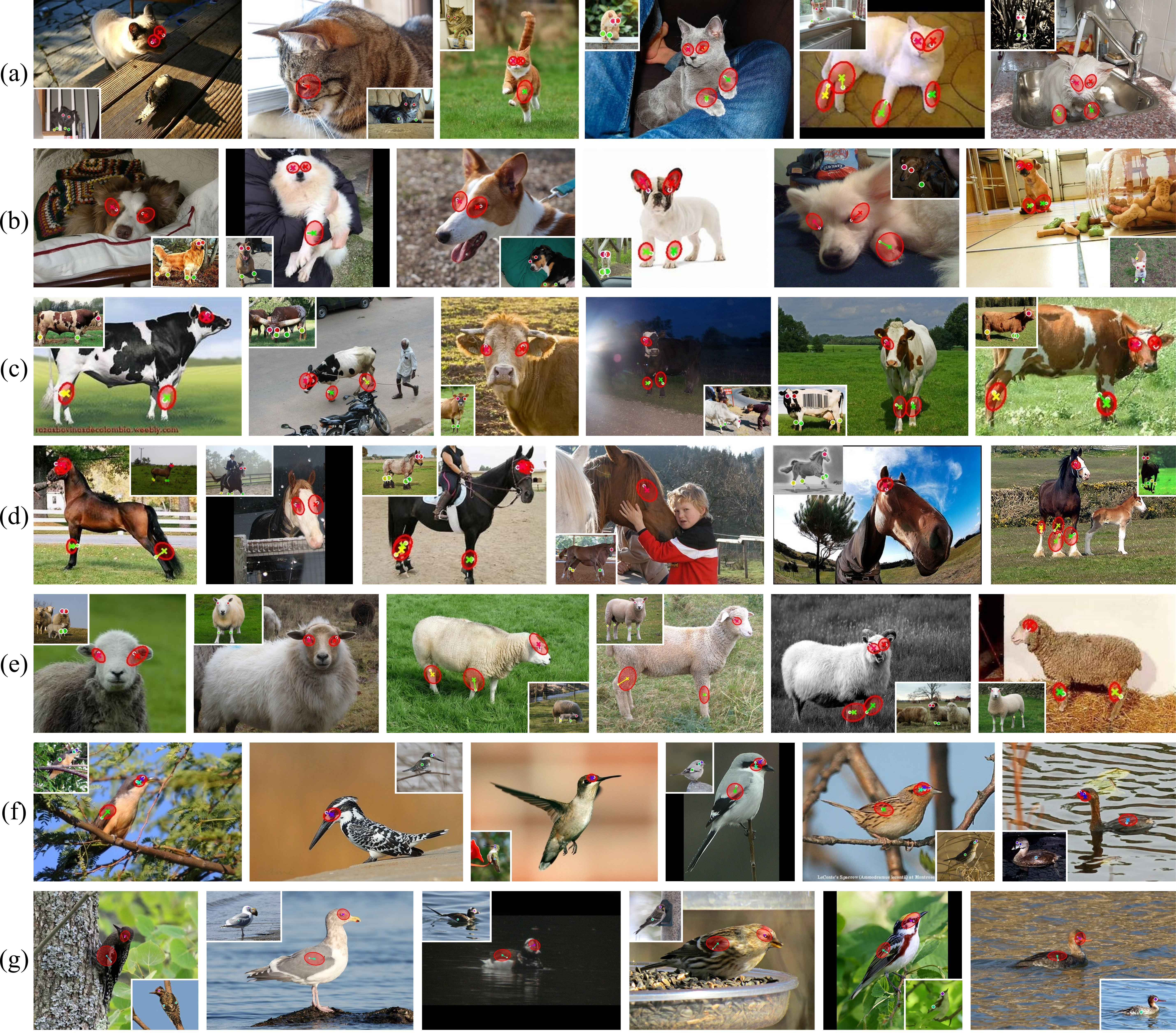}
  \caption{Extensive examples of 1-shot detection for novel keypoints in unseen species. From (a)$\sim$(e), each row is a subproblem by regarding an animal as unseen species in animal pose dataset, which is cat, dog, cow, horse, and sheep; (f) and (g) are results from 1-shot tasks in CUB and NABird, respectively. The experiments run in same-species episodes. The novel keypoint predictions (tilted crosses), estimated localization uncertainty (red ellipses), and groundtruth keypoints (circles) are simultaneously drawn.
  }
  \label{fig:fskd-visual2}
\end{figure*}
Moreover, we visualize additional 1-shot detection results for novel keypoints in unseen species. As shown in Fig. \ref{fig:fskd-visual2}, despite the query images containing various detrimental factors such as numerous behaviors, complex natural backgrounds, shadows, and areas of low contrast, the proposed FSKD  successfully detects novel keypoints in each query image given the support keypoints. Further, the estimated uncertainty marked by the red ellipse covers both the keypoint prediction and GT location, which indicates that the localization uncertainty is a good indicator of where the possible GT keypoint is located. Interestingly, the uncertainty distribution exhibits a relationship with the shape of body parts, which 
should help limit the ambiguity of keypoints. 

\section{Additional Ablation Study}
In this section, we present more ablation studies to validate the effectiveness of components involved in our pipeline. Similar to Section \ref{subsec:ablation-study} in the paper, we use the all-way-1-shot novel keypoints detection in unseen species with FSKD (default) running on same-species episodes.

\begin{figure}[!tb]
  \centering
  \includegraphics[width=\linewidth]{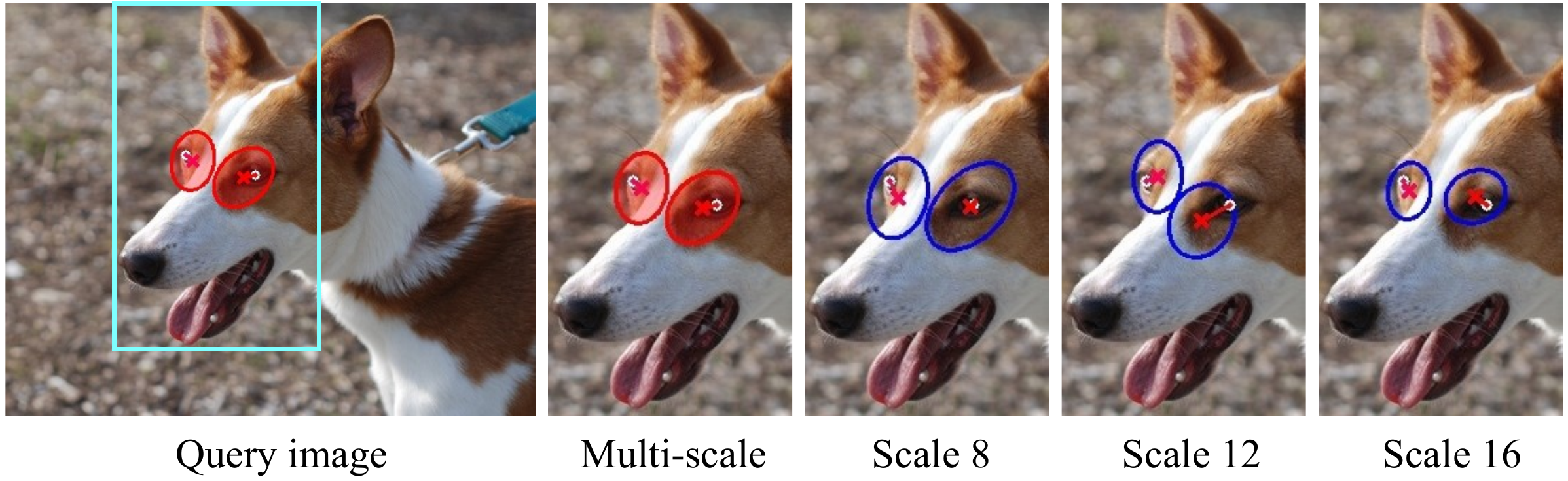}
  \caption{MS UC-GBL decomposition and uncertainty fusion.
  }
  \label{fig:multi-scale}
\end{figure}

\noindent\textbf{Additional Results on Multi-scale UC-GBL:}
We visualize outputs of each scale from Multi-scale UC-GBL, with an example shown in Fig.~\ref{fig:multi-scale}. Indeed, the keypoint prediction from multi-scale UC-GBL is more stable, reducing the risk of mislocalization. Increasing scale $S$ makes the grid finer and thus the uncertainty range shrinks. However, our fused uncertainty yields a good quality of combined uncertainty estimation.

\comment{
We visualize each-scale output from Multi-scale UC-GBL and an example is shown in Fig.~\ref{fig:multi-scale}. Indeed, the keypoint prediction from multi-scale is stabler and overall reduces the risk of mislocalization. When increasing scale $S$, the grid will be finer and thus the uncertainty will shrink. However, our fused uncertainty can give the proper ensemble estimation.
}
\begin{table}[!tb]
  \centering
  \newcommand{\tabincell}[2]{\begin{tabular}{@{}#1@{}}#2\end{tabular}}
  \caption{Study on keypoint feature extraction strategies and improvements by self-modulation during meta-testing. 
  }
  \small
  \label{tab:extraction-methods}
  \begin{tabular}{lccc}
      \toprule[1pt]
      \multirow{1}*{Extraction method} & Self-modulation& \multicolumn{1}{c}{Cat}  & \multicolumn{1}{c}{Dog} \\ \midrule[1pt]
        Integer-based indexing & \XSolidBrush           &48.11          &40.60    \\
        Bilinear interpolation & \XSolidBrush           &52.30          &47.18    \\
        Gaussian pooling       & \XSolidBrush           &\textbf{52.36} &\textbf{47.94}    \\ \midrule
        Gaussian pooling       & \emph{1 gradient-step} &52.88          &48.96    \\
        Gaussian pooling       & \emph{2 gradient-step} &53.66          &49.08    \\
        Gaussian pooling       & \emph{3 gradient-step} &54.01          &\textbf{49.40}    \\
        Gaussian pooling       & \emph{4 gradient-step} &\textbf{54.43} &49.00    \\
        Gaussian pooling       & \emph{5 gradient-step} &52.87          &49.36    \\
      \bottomrule[1pt]
    \end{tabular}
\end{table}

\noindent\textbf{Body Part Extraction Strategies:}
%
We compare the impacts of three keypoint feature extraction approaches such as the integer-based indexing, bilinear interpolation, and Gaussian pooling, given the same architecture, and 
perform experiments on two subproblems by regarding cat and dog as unseen species, respectively. Table \ref{tab:extraction-methods} shows that the bilinear interpolation and Gaussian pooling yield better results as the extracted soft keypoint representations contain larger spatial context, which helps build a more expressive support keypoint prototype (SKP).

\noindent\textbf{Self-modulation in Meta-testing:}
In addition, we find that the learnt FSKD model can improve its performance via self-modulation during meta-testing. Following Model-Agnostic Meta-Learning (MAML) \cite{finn2017model}, in each episode, we fine-tune the learnt meta-model via several gradient descent steps of back-propagation such that the meta-model has a chance to adapt better to the test data. Specifically, given the access to the support keypoints in the support image during meta-testing, we use 
SKPs to modulate the \emph{support feature map} and construct the loss using \emph{support keypoints}. After several gradient descent steps of back-propagation, the fine-tuned model is used for detecting the corresponding keypoints in the query image by modulating the \emph{query feature map}.  Table \ref{tab:extraction-methods} shows significant gains when using self-modulation, \eg, 54.43\% (4 gradient-step fine-tuning) \vs 52.36\% (without fine-tuning) for the cat. Meanwhile, we note that the excessive fine-tuning may overfit.

\comment{
In addition, we find that the learned FSKD model could improve its performance via self-modulation during meta-testing. Following Model-Agnostic Meta-Learning (MAML) \cite{finn2017model}, in each episode, we finetune the learned meta-model via several gradient steps of back-propagation such that the meta-model could adapt better. Specifically, as we have access to the support keypoints in support image during meta-testing, we use the built SKPs to modulate the \emph{support feature map} and construct the loss using support keypoints. After several gradient steps of back-propagation, the finetuned model is used to detect the corresponding keypoints in query image by modulating on \emph{query feature map}. As shown in Table \ref{tab:extraction-methods}, we can observe significant gains when using self-modulation, \eg, 54.43\% (4 gradient-step finetuning) \vs 52.36\% (without finetuning) in cat. Meanwhile, it should be careful to overfit and avoid finetuning too many gradient steps.
}

\begin{figure}[!tb]
  \centering
  \includegraphics[width=\linewidth]{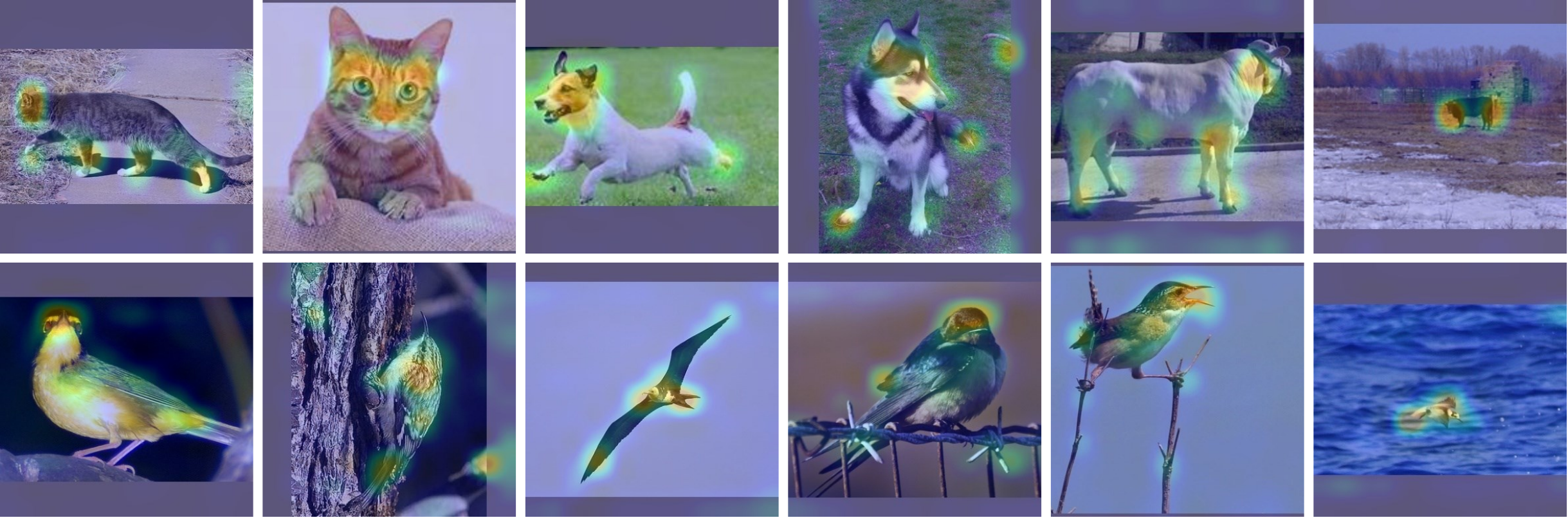}
  \caption{Visualization of semantic distinctiveness map $\boldsymbol{\sigma}^{-1}$.}
  \label{fig:visual-distinctiveness}
\end{figure}

\noindent\textbf{Semantic Distinctiveness Map:}
Examples of semantic distinctiveness (SD) map $\boldsymbol{\sigma}^{-1}$ are shown in Fig.~\ref{fig:visual-distinctiveness}.

\noindent\textbf{Mix-species Episode:}
To investigate the few-shot keypoint detection with the mix-species episodes, we perform the experiments on CUB and NABird with the goal of detecting keypoints in unseen species. Table~\ref{tab:kps-on-unseen-species-mix-episode} shows that the proposed FSKD variants are still effective but incur slight performance drops compared to the results in keypoint detection using same-species episodes, \eg, 51.22\% (Table~\ref{tab:kps-on-unseen-species-mix-episode}) \vs 56.04\% (Main paper, Table~\ref{tab:kps-on-unseen-seen-species}) achieved by FSKD (default) in NABird. The mix-species episode leads to a larger domain shift between the support and query images and thus poses an additional challenge to the model learning and keypoint localization.


\begin{table}[!tb]
  \centering
  \newcommand{\tabincell}[2]{\begin{tabular}{@{}#1@{}}#2\end{tabular}}
  \caption{Results on 1-shot keypoint detection for unseen species. The mix-species episode is used.
  }
  \small
  \label{tab:kps-on-unseen-species-mix-episode}
  \begin{tabular}{cccc}
      \toprule[1pt]
      \multirow{1}*{Setting} & \multirow{1}*{Method}  & \multirow{1}*{CUB} & \multirow{1}*{NABird} \\\midrule[1pt]
        \multirow{3}*{Novel}& Baseline                & 65.45              & 34.48                 \\
                            & ProbIntr                & 59.07              & 35.06                 \\
                            & FSKD (rand)             & 75.27              & 50.25                 \\
                            & FSKD (default)          & \textbf{76.99}     & \textbf{51.22}        \\\midrule
        \multirow{3}*{Base} & Baseline                & 80.81              & 74.10                 \\
                            & ProbIntr                & 71.40              & 74.82                   \\
                            & FSKD (rand)             & 86.92              & 80.25                   \\
                            & FSKD (default)          & \textbf{87.66}              & \textbf{84.74} \\
      \bottomrule[1pt]
    \end{tabular}
\end{table}


\section{Visualizations of Semantic Alignment}
\begin{figure*}[!t]
  \centering
  \includegraphics[width=.95\linewidth]{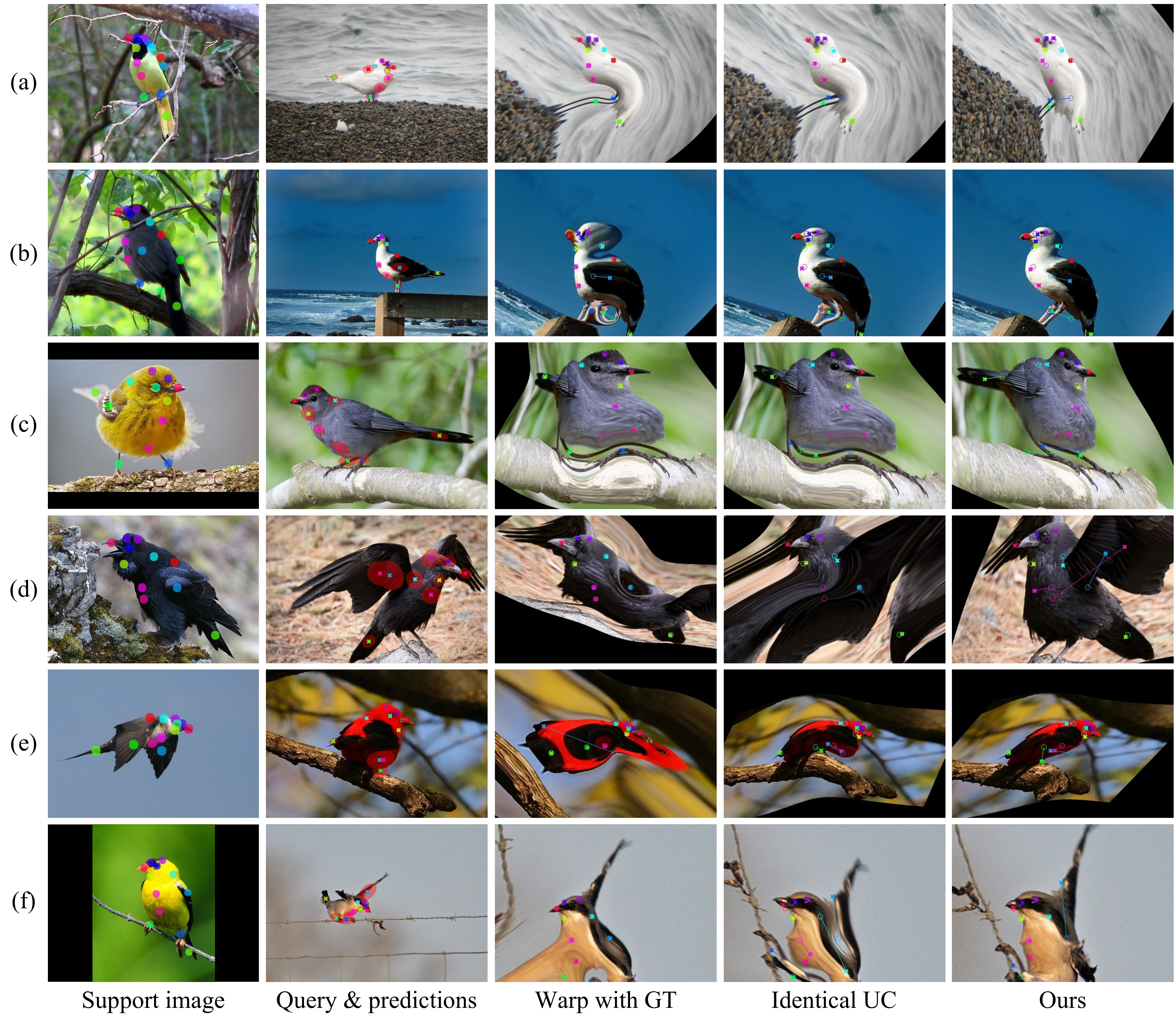}
  \caption{Additional qualitative results of semantic alignment using different approaches. The first column shows the support keypoints \& image; the second column shows the query image with the predicted keypoints (marked by tilted crosses) and uncertainty (red shadow ellipses); the last three columns are the results achieved by \emph{Warp with GT} \cite{bookstein1989principal}, \emph{Identical UC}, and our uncertainty-weighted TPS warp.}
  \label{fig:uc-tps-warp2}
  \vspace{-10pt}
\end{figure*}

Extensive qualitative results of semantic alignment (SA) are shown in Fig.~\ref{fig:uc-tps-warp2}. We perform SA for unseen species using 1-shot FSKD model trained on mix-species episodes. The reason we chose the mix-species episode setting is that aligning objects of different visual categories yields more diverse SA results in this setting.

When performing \emph{Warp with GT} query keypoints (Fig.~\ref{fig:uc-tps-warp2}, 3rd column), even though most query keypoints (marked by tilted crosses) align perfectly with the support keypoints (marked by circles), \emph{Warp with GT} results in unacceptable deformations of objects. In contrast, \emph{Identical UC} (Fig.~\ref{fig:uc-tps-warp2}, 4th column)  maintains the shape relatively better compared to \emph{Warp with GT} by applying the identical warping penalty. However, as weights of warping penalization are equal across keypoints, one can see that the deformations may appear in the proximity of inaccurate or poorly corresponding keypoints. In contrast to \emph{Warp with GT} and \emph{Identical UC}, our uncertainty-weighted TPS warp addresses the above issues, thus producing a much better perceptual alignment. Additionally, despite the support and query images are from different species and often have very differed poses, our FSKD detects the keypoints reliably, estimates the uncertainty reliably, and thus leads to a high-quality semantic alignment.


\section{Discussion}
\noindent\textbf{Difference Compared with Other Few-shot Tasks:}
Compared to few-shot image classification (FSL) and few-shot object detection (FSOD), there are \emph{two} main difference in FSKD.

Firstly, in FSL and FSOD, \emph{$N$-way learning} refers to $N$ visual categories in support set, while in FSKD, $N$-way means there are $N$ different keypoint types.  
Secondly, the \emph{training \& testing splits} are devised differently in FSKD, as detailed below.

Denote the set of classes of the training species  as $C=\{c_i\}_{i=1,2,\cdots,N_C}$ and the set of classes of testing species as $C'\!=\{c'_i\}_{i=1,2,\cdots,N_{C'}}$, where each element represents a class label. Let the set of training keypoint types be $\mathcal{X}=\{{k_i}\}_{i=1,2,\cdots,N_{\mathcal{X}}}$ and the set of testing keypoint types be $\mathcal{X}'=\{{k'_i}\}_{i=1,2,\cdots,N_{\mathcal{X}'}}$. 

In FSL and FSOD, one splits the species into base and novel class to guarantee $C \cap C'=\varnothing$. While in FSKD, one needs to \emph{split both species and keypoint types} and 
consider the following possible settings:
\begin{itemize}
\item $C \cap C'=\varnothing \text{ and } \mathcal{X} \cap \mathcal{X}'=\varnothing,$ $\;$(the hardest setting)
\item $C = C'\text{ and } \mathcal{X} \cap \mathcal{X}'=\varnothing$, $\quad$(intermediate difficulty)
\item $C \cap C'=\varnothing \text{ and } \mathcal{X} = \mathcal{X}'$, $\quad$(intermediate difficulty)
\item $C = C' \text{ and } \mathcal{X} = \mathcal{X}'$. $\qquad\qquad$(the easiest setting)
\end{itemize}


\noindent\textbf{Limitations and the Future Work:}
Learning from several annotated samples to detect novel keypoints is hard but it could be improved with  more expressive keypoint representations beyond the Gaussian pooling and advanced feature modulation schemes. Furthermore, the auxiliary keypoints interpolated along lines are suboptimal due to their imprecise matching relationship in locations between support and query, which yields a relatively large localization noise. Despite we address the impact of noise via our uncertainty modeling, we hope to improve the signal-to-noise ratio with more advanced interpolation strategies that could take each shape of object into account.



\comment{
1. The auxiliary keypoints may accidentally pass through novel keypoints, and thus the FSKD may not strictly detect the novel keypoints as they may be seen during training phase.
Answer: The authors acknowledge that the auxiliary keypoints may have chance to pass through novel keypoints. However, the probability is very small as the auxiliary keypoints are automatically interpolated from body parts. Basically, FSKD is a very hard problem due to its large domain shift in settings (here we need to detail the current easiest setting in current works, hardest setting), keypoint ambiguity/without clear boundary/refer to local region, and large existence of noise, which resists the system to detect novel keypoints. Due to the difficulty of FSKD, therefore, we have to use auxiliary keypoints to bridge the gap and reduce the domain shift. Moreover, the training keypoint types from annotations are quite limited, the usage of auxiliary keypoints could greatly increase the visual diversity of keypoints, which may help FSKD model infer the novel keypoints. At last, the auxiliary keypoints didn't involve human priors and are automatically generated (akin to self-supervision), thus it is reasonable to be added into training.
}


\bibliographystylelatex{ieee_fullname}
\bibliographylatex{refs_supp}

\end{document}